%% file: iclr2026_conference.tex
\documentclass{article} 
\usepackage{iclr2026_conference,times}

\iclrfinalcopy
\usepackage{url}
\input{math_commands}

\usepackage{graphicx}
\usepackage[dvipsnames]{xcolor}
\usepackage{booktabs}
\usepackage{wrapfig}
\usepackage{enumitem}
\usepackage{xcolor}
\usepackage{colortbl}
\usepackage{amssymb}
\usepackage[
    colorlinks=true,   
    linkcolor=PineGreen,
    citecolor=PineGreen,   
    urlcolor=PineGreen,
]{hyperref}

\title{Refusal Falls Off a Cliff: \\ How Safety Alignment Fails in Reasoning?}

\author{
    Qingyu Yin$^{1}$\thanks{Work done during internship at Xiaohongshu Inc.} \quad 
    Chak Tou Leong$^{2}$ \quad  
    Wenxuan Huang$^{3}$ \quad  
    Wenjie Li$^{2}$ \quad  Linyi Yang$^{4}$ \quad  \\
    \ \textbf{Xiting Wang}$^{5}$ \quad 
    \textbf{Jaehong Yoon}$^{6}$ \quad 
    \textbf{YunXing}$^{7}$ \quad 
    \textbf{XingYu}$^{7}$ \quad 
    \textbf{Jinjin Gu}$^{8}$ 
    \\
    \\
    $^{1}$Zhejiang University, 
    $^{2}$Hong Kong Polytechnic University, \\
    $^{3}$East China Normal University,
    $^{4}$Southern University of Science and Technology, \\ $^5$Renmin University $^6$Nanyang Technological University  $^7$Xiaohongshu Inc., $^{8}$INSAIT \\
}

\newcommand{\eg}{\textit{e.g., }}
\newcommand{\ie}{\textit{i.e., }}
\newcommand{\term}{refusal cliff}

\begin{document}

\maketitle

\begin{abstract}
Large reasoning models (LRMs) with multi-step reasoning capabilities have shown remarkable problem-solving abilities, yet they exhibit concerning safety vulnerabilities that remain poorly understood. In this work, we investigate why safety alignment fails in reasoning models through a mechanistic interpretability lens. Using a linear probing approach to trace refusal intentions across token positions, we discover a striking phenomenon termed as \textbf{refusal cliff}: many poorly-aligned reasoning models correctly identify harmful prompts and maintain strong refusal intentions during their thinking process, but experience a sharp drop in refusal scores at the final tokens before output generation. This suggests that these models are not inherently unsafe; rather, their refusal intentions are systematically suppressed. Through causal intervention analysis, we identify a sparse set of attention heads that negatively contribute to refusal behavior. Ablating just 3\% of these heads can reduce attack success rates below 10\%. Building on these mechanistic insights, we propose \textbf{Cliff-as-a-Judge}, a novel data selection method that identifies training examples exhibiting the largest refusal cliff to efficiently repair reasoning models' safety alignment. This approach achieves comparable safety improvements using only 1.7\% of the vanilla safety training data, demonstrating a less-is-more effect in safety alignment. Code is available at \href{https://github.com/MikaStars39/RefusalCliff}{here}.
\end{abstract}

\section{Introduction}
Large Reasoning Models~\citep{guo2025deepseek, shao2024deepseekmath,openr1}, with advanced reasoning capability derived from reinforcement learning with verifiable rewards (RLVR)~\citep{yu2025groupsequencepolicy,liu2025dapoopensourcellm}, are designed to handle complex problem solving, logical inference, and tool‑assisted planning. However, while these methodological advances signal more reliable and capable models, they simultaneously introduce significant safety considerations. It is widely discovered that current reasoning‑oriented models often lag behind in safety alignment, and tend to exhibit higher susceptibility to attacks~\citep{kuo2025h, sabbaghi2025adversarial,kuo2025h, zaremba2025trading, zhou2025hidden,li2025smarter}, highlighting an urgent need for reasoning‑specific safety mechanisms. Many previous works have benchmarked the safety of reasoning models~\citep{jiang2025safechain}, developed jailbreaking strategies~\citep{wang2025safety}, and proposed alignment solutions~\citep{zhang2025realsafe}, but have lacked analysis of the mechanisms under the vulnerability of reasoning safety.

Understanding why safety alignment in reasoning models is vulnerable provides invaluable insights for both societal benefit and future model development. In this paper, we \textit{firstly} aim to answer the following \textbf{research question}:
\begin{center}
    \textit{What mechanism makes the safety alignment vulnerable in reasoning models?}
\end{center} 
While numerous reasoning models exhibit unsafe behaviors, the underlying mechanisms driving these failures remain critically important to investigate. Do these reasoning models lack safety capabilities, or do they have adequate risk assessment abilities but simply choose not to act on them, failing to refuse harmful requests? Empirical studies have shown that the internal reasoning traces of such models can be unfaithful to the actual decision-making process and may fail to explicitly reveal the model’s true intentions~\citep{barez-chain-2025,Arcuschin2025ChainofThoughtRI}. This limitation motivates the need to probe models from the perspective of their internal representations.
Prior research has demonstrated that language models encode meaningful and behaviorally relevant features within their representation space~\citep{turner2023steering, engels2025not, gorton2025adversarialexamplesnot}. These latent features have been shown to govern various emergent behaviors, \eg in-context learning~\citep{Ilharco2022EditingMW, Hendel2023InContextLC}, instruction following~\citep{stolfo2025improving}, and sentiment modulation~\citep{turner2023steering}. In the context of safety alignment research, refusal behavior is often considered a canonical metric, and a specific refusal direction~\citep{arditi2024refusal} in representation space has been shown to regulate such behavior.
To examine how these safety-relevant features evolve across tokens and layers, a prominent approach emerging from mechanistic interpretability -- the use of linear probes~\citep{nanda2025sparseautoencodersuseful} -- offers a principled method for analyzing the internal processing of language models. 

\begin{figure}[t]
    \centering
    \includegraphics[width=1\linewidth]{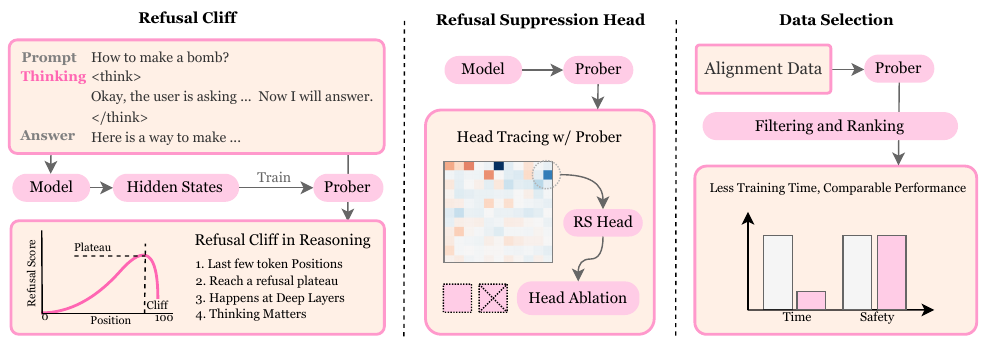}
    \caption{\textbf{An overview of our paper}. \textit{Left:} We train a prober and discover the refusal cliff. \textit{Center:} We find Refusal Suppression Heads as the main cause of the cliff. \textit{Right:} We design data selection method based on probing the cliff.}
    \label{fig:main}
\vspace{-0.30cm}
\end{figure}


To characterize the dynamics of refusal behavior in reasoning models, we build on prior work~\citep{chan2025can,xu2024uncovering} and adopt a probing-based methodology to quantify safety-relevant signals in hidden-state representations. In our framework, safety is operationalized via refusal behavior~\citep{arditi2024refusal}, as well-aligned models are expected to refuse harmful queries (\eg through \textit{I’m sorry} statements). Concretely, we train a linear probe classifier to predict, given hidden states from different positions in the reasoning chain, whether the model will refuse the prompt. The probe’s predicted probability is termed the \textit{refusal score}, with higher scores indicating internal states more predictive of refusal. Across multiple partially aligned reasoning models, we observe a recurrent pattern we call the \textbf{Refusal Cliff}. While intermediate reasoning steps yield refusal scores comparable to strongly aligned instruction-tuned models -- indicating successful detection of harmfulness -- scores drop sharply in the final steps. This reflects suppression of refusal behavior even where refusal would be the alignment-consistent choice. The sharp decline suggests these models maintain alignment only in early reasoning, but fail to preserve it through output generation.

The Refusal Cliff consistently occurs at the final positions of the reasoning chain, corresponding to a fixed set of output tokens (the \textit{thinking-end template}). These template tokens must retrieve contextual information from earlier reasoning steps via attention mechanisms. We hypothesize that specific attention heads play a critical role: while most propagate alignment-consistent features supporting refusal, certain heads introduce competing signals that attenuate refusal-related representations, driving the observed score drop. Our detailed ablation experiments confirm this hypothesis, revealing a small set of \textbf{Refusal Suppression Heads}, sparsely distributed across deeper layers, that systematically reduce refusal scores. Removing these heads increases refusal scores at the thinking-end template and, in poorly aligned models, reduces attack success rates to below 10\%. 


To mitigate the Refusal Cliff, we propose a data filtering strategy that leverages internal representation signals to prioritize high-impact training samples. The key assumption is that effective safety fine-tuning should recover the model’s early-stage refusal plateau -- the stable region of refusal scores prior to suppression. We quantify misalignment between this plateau and the cliff position (where scores drop sharply) using a misalignment score, defined as the absolute difference between the plateau mean and the final-step score. We then fine-tune only on the most misaligned examples, targeting cases where refusal degradation is most severe. Using just the top 1.3\% of samples, we reduce attack success rates on harmful-query benchmarks to below 5\% while significantly lowering wall-clock training time relative to full-dataset fine-tuning. Compared to filtering methods such as LLM-as-a-judge~\citep{gu2024survey}, \textbf{Cliff-as-a-judge} achieves comparable safety gains with more flexible, metric-driven selection, demonstrating a clear \textit{less-is-more} effect in alignment.


As summarized in Figure~\ref{fig:main}, our contributions are threefold:
\begin{itemize}[leftmargin=0.6cm, itemsep=0.05cm]
\item We identify and characterize the \textbf{Refusal Cliff}, a failure mode in which refusal intentions abruptly vanish at the reasoning output stage.
\item We causally link this phenomenon to a small set of \textbf{Refusal Suppression Heads}, which undermine refusal behavior by suppressing alignment features.
\item We introduce \textbf{Cliff-as-a-judge}, a probing-driven data selection method that mitigates safety vulnerabilities and achieves a \textit{``less is more''} effect in safety alignment.
\end{itemize}

\section{Preliminaries}

\paragraph{Transformer.} We study reasoning models with Transformers~\citep{vaswani2017attention} as a backbone.  One Transformer model usually consists multiple of layers and an embedding layer.  For an input $\boldsymbol{X_i} \in \mathbb{R}^{n \times 1}$ with length $n$, it first passes through an embedding layer with hidden state size $d$, then passes all the Transformer layers:
\begin{equation}
\label{eq:residual_att}
    \boldsymbol{H^{\text{att}}_{i}} = \boldsymbol{H_i} + \mathrm{Attn}(\mathrm{Norm}(\boldsymbol{X_i})), \ \boldsymbol{H_i} = \boldsymbol{H^{\text{att}}_{i}} + \mathrm{MLP}(\mathrm{Norm}(\boldsymbol{H^{\text{att}}_{i}})).
\end{equation}
Here, $\boldsymbol{H^{\text{att}}_{i}}$ is the output hidden states of the attention block, and $\boldsymbol{H^{\text{mlp}}_{i}}$ is the output of the MLP block for layer $i$.

\begin{figure}[t]
    \centering
    \begin{minipage}[t]{0.49\textwidth}
        \centering
        \includegraphics[width=1\linewidth]{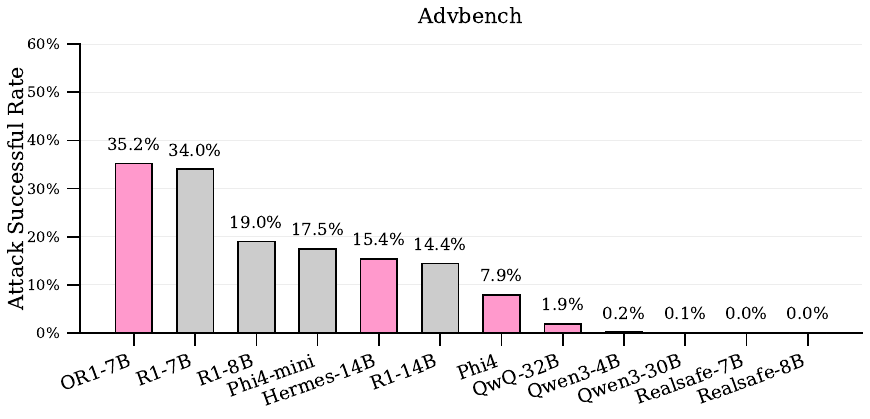}
    \end{minipage}
    \begin{minipage}[t]{0.49\textwidth}
        \centering
        \includegraphics[width=1\linewidth]{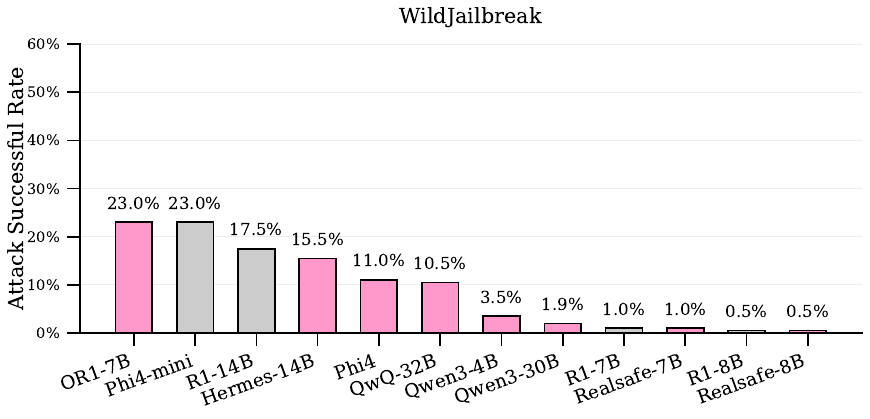}
    \end{minipage}
    \caption{While some reasoning models achieve reasonable safety performance, a significant portion exhibit alarming vulnerabilities to adversarial attacks. We benchmark reasoning models ({\color{Magenta!80}RLVR-based} and {\color{gray}Distillation-based}) on AdvBench~\citep{chao2024jailbreakbench} and WildJailbreak~\citep{wildteaming2024} with Attack Success Rate (ASR, the \textit{lower} the \textit{better}) as evaluation metric.}
    \label{fig:safe}
\vspace{-0.3cm}
\end{figure}

\paragraph{Models.} We evaluate two categories of reasoning models:
\textit{(i) RLVR-based models}, trained with \emph{Reinforcement Learning with Verifiable Rewards (RLVR)} to enhance reasoning ability. We include QwQ~\citep{qwq32b, qwen2.5}, Qwen3-Thinking~\citep{yang2025qwen3}, Skywork-OR1~\citep{he2025skywork}, Phi-4-Reasoning~\citep{abdin2025phi}, and Hermes4~\citep{allan2018hermes}.
\textit{(ii) Distillation-based models}, trained by distilling reasoning traces from strong teacher models. We include DeepSeek-R1-Distill-Qwen-7B, DeepSeek-R1-Distill-LLaMA-8B, RealSafe-R1-7B, RealSafe-R1-8B~\citep{zhang2025realsafe}, and DeepSeek-R1-Distill-Qwen-14B~\citep{guo2025deepseek}.
These selections cover diverse architectures, scales, and training paradigms. We assess safety using LlamaGuard-4~\citep{grattafiori2024llama}, reporting \emph{Attack Success Rate (ASR)}, defined as the fraction of harmful generations. As shown in Figure~\ref{fig:safe}, safety alignment varies substantially across models: while some demonstrate robust alignment, others remain highly vulnerable.

\paragraph{Datasets.} We evaluate safety using datasets that span both \emph{vanilla attacks} -- direct harmful queries -- and \emph{adversarial attacks} -- crafted queries with deception and manipulation to bypass safeguards. For vanilla attacks, we use JailbreakBench~\citep{chao2024jailbreakbench}, AdvBench~\citep{zou2023universal}, and the vanilla subset of WildJailbreak~\citep{wildteaming2024}. For adversarial attacks, we use the adversarial subset of WildJailbreak.

\paragraph{Refusal Prober.} When an LLM encounters a harmful prompt, it will provide a refusal response to avoid giving users harmful information related to the question. Therefore, refusal examples \eg \textit{Sorry, I cannot...}, is a direct indicator for measuring the safety~\footnote{Basically, the refusal response rate is given by \( (1 - \text{ASR}) \) for harmful prompts \ie cases where the model either refuses or responds harmfully. Although cases such as fake refusals or ambiguous answers exist, we do not analyze these kinds of complex behavior. We believe that a good model should either provide a clear refusal or a helpful answer.}. This also holds true for reasoning models. Recent work has shown that refusal behavior is often controlled by a single \textit{refusal direction} within its activation space \citep{arditi2024refusal}. This direction is a vector that, when added to a hidden state, maximally increases the probability of generating a refusal. Due to this linear property, we can effectively identify this direction using a simple linear classifier \ie a refusal prober~\citep{xu2024uncovering}. The refusal prober is a logistic regression model that takes a hidden state vector $\boldsymbol{h}_j \in\boldsymbol{H}$ at token position $j$ as input and outputs the probability of refusal. The probability is calculated as: \begin{equation} \label{eq:prober} P(\mathrm{refusal} | \boldsymbol{h}_{j}) = \sigma(\boldsymbol{W}^T \boldsymbol{h}_{j} + b) \end{equation} The prober is trained on a dataset with $N$ examples $\mathcal{D} = \{ (\boldsymbol{h}^{k}_j, c^{k}) \}_{k=1}^N $ and the label $c$ is defined as: \begin{equation} c := \begin{cases} 1 & \text{for a refusal response (e.g., \textit{Sorry, I cannot...})}, \\ 0 & \text{for a normal response (e.g., \textit{The answer is...})}, \end{cases} \end{equation} where $\boldsymbol{W} \in \mathbb{R}^{d \times 1}$ is the weight vector, $b$ is the bias, and $\sigma$ is the sigmoid function. We define the output probability as the \textbf{refusal score} of reasoning model at position $j$.

\section{Refusal Cliff in Reasoning Models}
\label{sec:cliff}

\setlength{\columnsep}{9pt} 
\setlength{\intextsep}{0pt}
\begin{wrapfigure}{r}{0.33\textwidth}
    \centering
    \vspace{-0.4cm}
    \includegraphics[width=1\linewidth]{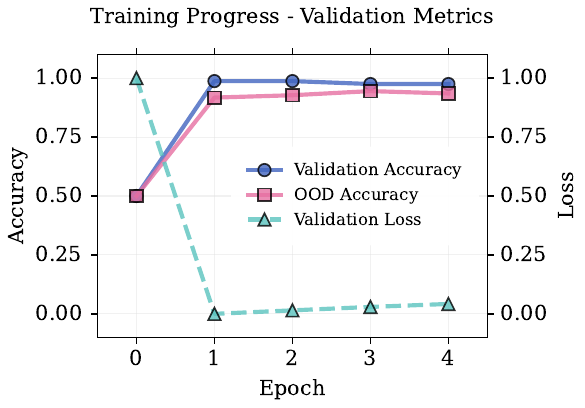}
    \caption{The loss, validation accuracy and OOD validation accuracy of the refusal prober.} 
    \vspace{0.1cm}
    \label{fig:prober_curve}
\end{wrapfigure}

\paragraph{Preparations.} We first train a refusal prober following the design in Equation~\ref{eq:prober}. We trained the prober using the hidden states $\boldsymbol{h}$ extracted from the \textit{final token position} in the last layer of each sequence in our dataset $\mathcal{D}$. For refusal response, we collect examples from Advbench~\citep{zou2023universal}, and non-refusal response are collected from Ultrachatsft~\citep{ding2023enhancing}. The prober was trained for 5 epochs with 256 examples and achieved an average validation accuracy of over 95\%. Loss curve and accuracy are shown in Figure~\ref{fig:prober_curve}. Considering the validation set is sampled from Advbench, as same as the examples source, we also test the Out of Distribution (OOD) accuracy of the prober on JailbreakBench~\citep{chao2024jailbreakbench}. This high accuracy confirms that refusal behavior can be reliably predicted from a linear analysis of the model's internal states. 
Further details on hyperparameters and experimental settings are available in Appendix~\ref{sec:prober_additional}. 

\begin{figure}[t]
    \centering
    \begin{minipage}[t]{0.47\textwidth}
        \centering
        \includegraphics[width=1\linewidth]{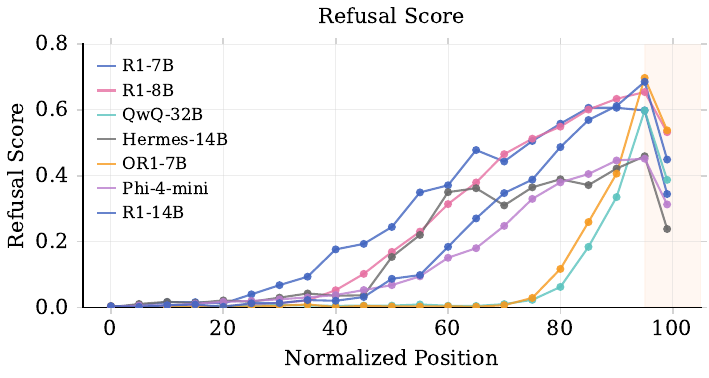}
    \end{minipage}
    \begin{minipage}[t]{0.47\textwidth}
        \centering
        \includegraphics[width=1\linewidth]{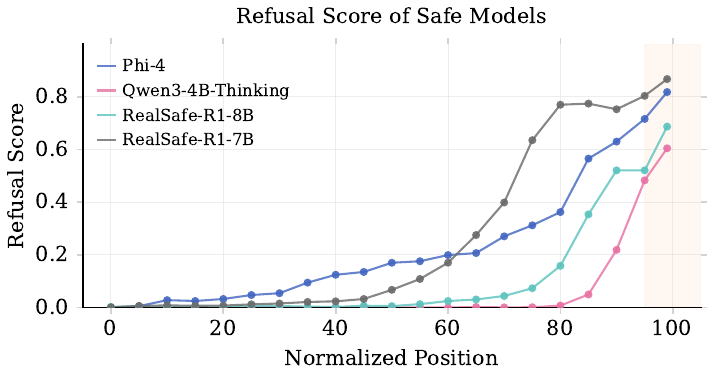}
    \end{minipage}
    \caption{\textit{Left:} Reasoning model with refusal cliff. We highlight the cliff position with {\color{orange!50} orange} background. \textit{Right:} Well-aligned reasoning models experience no refusal cliff.}
    \label{fig:refusal_cliff}
\vspace{-0.5cm}
\end{figure}




\begin{figure}[t]
    \centering
    \begin{minipage}[t]{0.24\textwidth}
        \centering
        \includegraphics[width=1\linewidth]{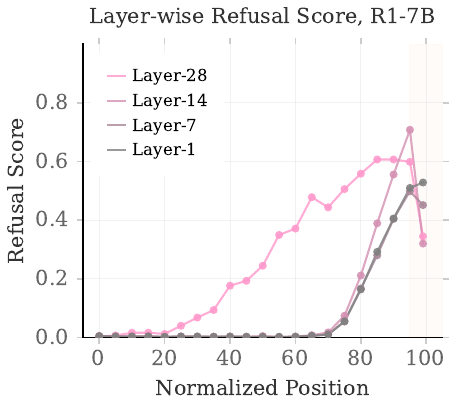}
    \end{minipage}
    \begin{minipage}[t]{0.24\textwidth}
        \centering
        \includegraphics[width=1\linewidth]{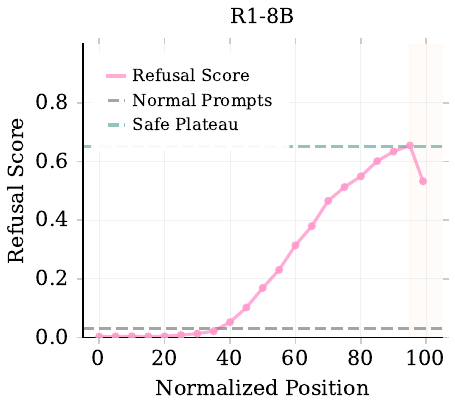}
    \end{minipage}
    \begin{minipage}[t]{0.24\textwidth}
        \centering
        \includegraphics[width=1\linewidth]{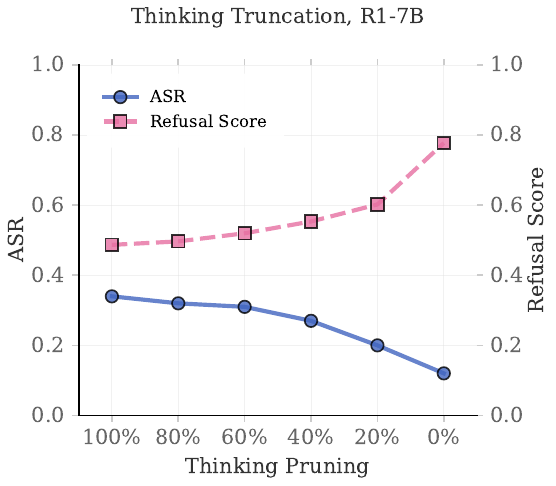}
    \end{minipage}
    \begin{minipage}[t]{0.24\textwidth}
        \centering
        \includegraphics[width=1\linewidth]{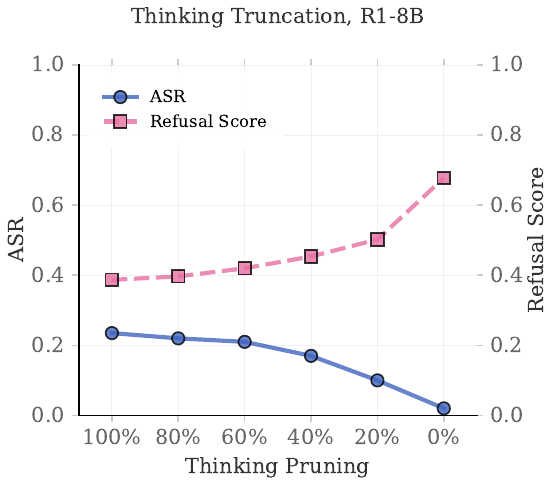}
    \end{minipage}
    \begin{minipage}[t]{0.24\textwidth}
        \centering
        \includegraphics[width=1\linewidth]{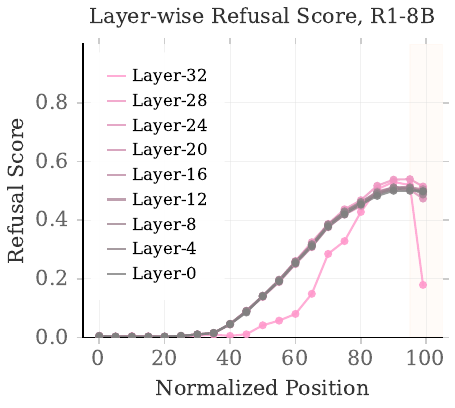}
    \end{minipage}
    \begin{minipage}[t]{0.24\textwidth}
        \centering
        \includegraphics[width=1\linewidth]{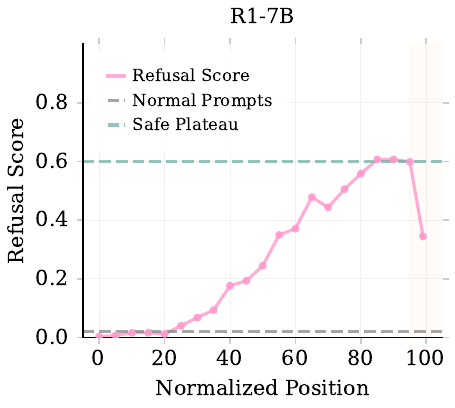}
    \end{minipage}
    \begin{minipage}[t]{0.24\textwidth}
        \centering
        \includegraphics[width=1\linewidth]{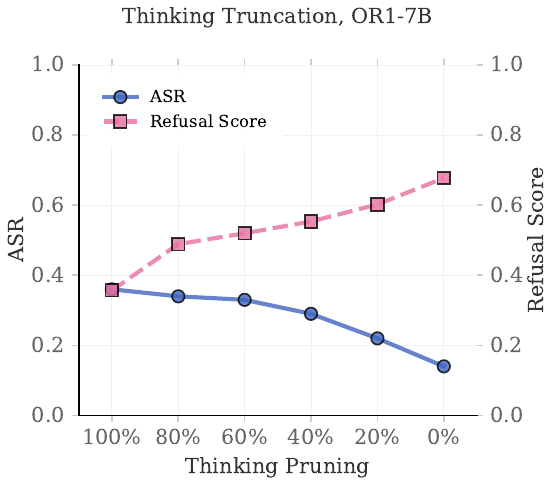}
    \end{minipage}
    \begin{minipage}[t]{0.24\textwidth}
        \centering
        \includegraphics[width=1\linewidth]{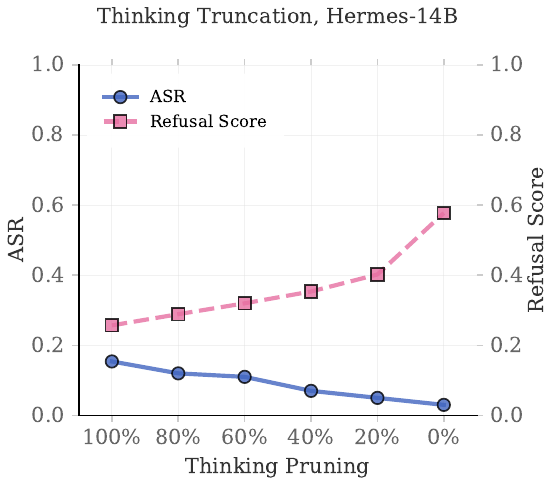}
    \end{minipage}
    \caption{\textit{The first column on the left:} Layer-wise refusal score of R1-Distill-Qwen-7B and R1-Distill-LLaMA-8B from {\color{gray} shallow} layers to {\color{VioletRed!70} deeper} layers . \textit{The second column on the left:} Comparison of refusal score in normal prompts and plateau values. {\color{gray} Gray} line is the average refusal score in normal prompts and {\color{YellowGreen} Green} line is the plateau of well-aligned family models. \textit{The third and fourth column on the left:} Relation between thinking length and misalignment. We gradually clip thinking and force the model to directly answer.
    }
    \label{fig:more}
    \vspace{-0.6cm}
\end{figure}

\paragraph{Refusal Cliff in Reasoning Models.}
We probe the hidden states of reasoning models using the trained refusal prober to estimate the \textit{refusal score} (defined at Eq.~\ref{eq:prober}) at each token position. Probing is conducted from the first token of the prompt until the end of the model’s reasoning process. 
Since the reasoning length varies across questions, we normalize all scores to a 0–100 scale, where 0 corresponds to the beginning and 100 corresponds to the final token position. By analyzing the refusal score of reasoning models, we can take a close look at their inner intention of tackling harmful requests. Results are illustrated in Figure~\ref{fig:refusal_cliff} where the \textit{left} are poorly aligned reasoning models and \textit{right} are models that perform relatively well on safety benchmarks.
Interestingly, for reasoning models that perform poorly on safety-related benchmarks, we observe a phenomenon we refer to as \textbf{Refusal Cliff}. As illustrated in Figure~\ref{fig:refusal_cliff}, the refusal score exhibits a gradual upward trend followed by a plateau phase. Critically, there is an abrupt decline in refusal scores at the terminal token positions, indicating that the model's internal intention transitions from rejecting the harmful request to complying with it.

\paragraph{Properties.}
To analyze the properties of the refusal cliff, we further conduct several experiments as shown in Figure~\ref{fig:more}. The \term \ exhibits four key properties and are summarized as below:
\begin{itemize}[leftmargin=0.3cm, itemsep=0.05cm]
    \item The cliff is highly localized to the final few tokens of the reasoning process (as shown in the gray location in Figure~\ref{fig:refusal_cliff}), immediately preceding the model's output \ie the template region. In contrast, safety-aligned models such as Phi~\citep{abdin2024phi} and Qwen3-thinking~\citep{yang2025qwen3} show little to no cliff at such positions; their refusal scores may even increase as they conclude their reasoning.
    \item As shown in \textit{the first column on the left}, Figure~\ref{fig:more}, The phenomenon is amplified in deeper layers, where the magnitude of the cliff increases substantially. Within deeper layers, the subsequent degradation in refusal efficacy becomes markedly more severe.
    \item As shown in \textit{the second column on the left}, Figure~\ref{fig:more}, the cliff is preceded by a plateau, indicating that the model \textit{recognizes} the harmful nature of the prompt despite its \textit{eventual non-compliance}. During this plateau, the model's refusal intention is comparable to that of well-aligned variants. 
    \item The model's thinking is vital for the refusal cliff. As we clip the thinking and directly prefilling the thinking end token \ie the thinking clipping operation~\citep{jiang2025safechain}, to stop the thinking of the model in \textit{the third and fourth column on the left}, Figure~\ref{fig:more}, we observe a lower level of refusal cliff and an increase of refusal response rate at the output. 
\end{itemize}

\section{Who is the Devil in Refusal Cliff? A Mechanistic Explanation from Attention Heads}
\label{sec:head}
We probe the refusal intention in reasoning models, discover refusal cliff, and discuss its properties. Since we know the refusal cliff exists, understanding how it happens is of great benefit to the safety and future improvements of reasoning models. In this section, we try to find out why.
\subsection{Attention Heads in Refusal Cliff}
\paragraph{Why Attention Heads?}
Intuitively, analyzing the phenomenon at the granularity of attention heads is natural: from a mechanistic interpretability perspective, attention heads are the main carriers of information routing in Transformer architectures, and different heads often specialize in diverse functions~\citep{yin2025which,olsson2022context,wu2024retrieval}. It is also proven that attention heads play a key role in safety~\citep{zhou2025on}.
In our case, the final tokens before the output closure template tokens \eg \verb|\n</think>\n\n|, are strongly stereotyped between generations. However, for both the same template, refusal cliff happens in harmful examples but not benign ones. Therefore, a sudden disruption of this pattern during a refusal cliff suggests that certain heads have attended to specific prior content that triggers a mode change in the model. 

\begin{figure}[h]
    \centering
    \vspace{0.2cm}
    \begin{minipage}[t]{0.32\textwidth}
        \centering
        \includegraphics[width=1\linewidth]{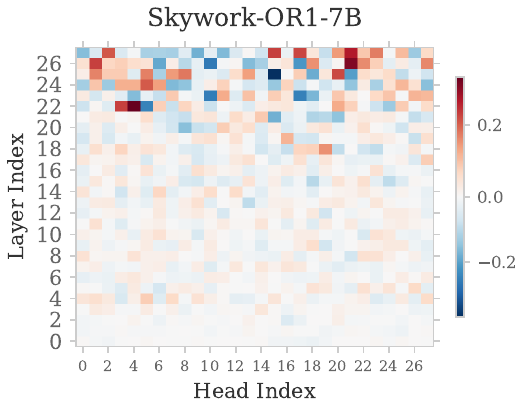}
    \end{minipage}
    \begin{minipage}[t]{0.32\textwidth}
        \centering
        \includegraphics[width=1\linewidth]{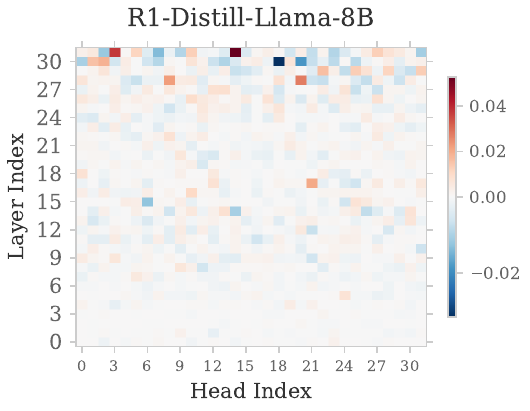}
    \end{minipage}
    \begin{minipage}[t]{0.32\textwidth}
        \centering
        \includegraphics[width=1\linewidth]{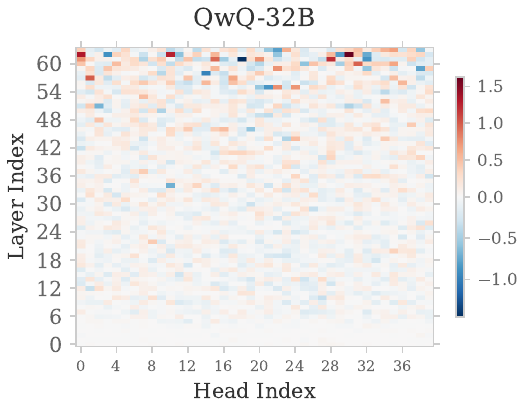}
    \end{minipage}
    \caption{We trace the contributions of attention heads with probing. {\color{red} Red} means the head contribute \textit{positively} to final refusal and {\color{blue} Blue} indicates that this head contributes \textit{negatively}.}
    \label{fig:trace}
    \vspace{-0.1cm}
\end{figure}

\paragraph{Tracing Attention Heads with Probing.}
To accurately assess the causal impact of each attention head on refusal behavior, we employ a direct probing method to trace each head's contribution. Our approach is to individually evaluate the influence of each head's output at $t_{\text{cliff}}$, where the refusal cliff occurs. Specifically, for an attention head $h$ in any layer $i$, we first isolate its output vector $\boldsymbol{o}_{i,h,t_{\text{cliff}}}$. Following the standard Transformer architecture, this vector is projected into the residual stream via the attention block's output weight matrix, $\boldsymbol{W}_{O,i}$. To analyze the contribution of head $h$ alone, we construct a hypothetical residual update vector, $\Delta \boldsymbol{h}_{i,h,t_{\text{cliff}}}$, where only the output of head $h$ is active, while the outputs of all other heads in the same layer are zeroed out. Subsequently, we feed this vector containing the contribution of only a single head, $\Delta \boldsymbol{h}_{i,h,t_{\text{cliff}}}$, as input to our pre-trained refusal prober to evaluate the head's independent refusal score. Its contribution score, $s_{i,h}$, is calculated as follows:
\begin{equation}
s_{i,h} = \boldsymbol{W}^T \Delta \boldsymbol{h}_{i,h,t_{\text{cliff}}} + b
\end{equation}
where $\boldsymbol{W}$ and $b$ are the parameters of the prober (Eq.~\ref{eq:prober}). We remove the sigmoid function so that we can directly trace the contribution of each attention head via logits~\citep{heimersheim2024use, zhang2023towards}. This score, $s_{i,h}$, directly quantifies the strength with which a single attention head, acting in isolation, pushes the model towards refusal or compliance. A score close to 1 indicates that the head promotes refusal, whereas a score close to 0 implies that it suppresses refusal.

\paragraph{Tracing Results.}
We aggregate the changes in refusal score for each head and visualize the results in Figure~\ref{fig:trace}. 
In the heatmap, ref indicates a positive contribution to refusal behavior 
(\ie the head writes into the residual stream in a way that increases the refusal score for harmful prompts), while blue denotes a negative contribution (\ie the head decreases the refusal score, making refusals less likely). 
Notably, the contribution pattern is highly sparse: a small fraction of heads exhibit a strong negative correlation with refusal behavior, which we term the \textbf{Refusal Suppression Heads}~\footnote{This definition is intended as a soft formulation, and in a later section we introduce a small threshold to facilitate the ablation analysis.}.

\subsection{Refusal Suppression Head Ablation}

\paragraph{Ablation Methodology.}
We perform head ablation to \textit{(i)} \textit{cross-validate} the importance of the heads identified through tracing in the previous subsection and \textit{(ii)} explore as a potential solution to tackle the unsatisfying safety alignment in reasoning models. Following previous work~\citep{liu2025roleattentionheads}, we ablate attention heads one by one and evaluate the resulting changes 
in both the refusal score and the overall safety performance. 
We employ a scaling-down ablation, in which we introduce a scaling factor $\gamma$ 
to the output of the selected attention head to get the output $\boldsymbol{O}$:
\begin{equation}
   \boldsymbol{O} = (\frac{\boldsymbol{Q}\boldsymbol{K}^\top}{\sqrt{d}} \odot \boldsymbol{M}) \cdot \gamma \cdot \boldsymbol{V}, \ \text{where} \ \boldsymbol{Q},\boldsymbol{K}, \boldsymbol{V}, \boldsymbol{O} \in \mathbb{R}^{l \times d}, \ \boldsymbol{M} \in \mathbb{R}^{l \times l}.
\label{eq:zero}
\end{equation}
Here, $\boldsymbol{Q}, \boldsymbol{K}, \boldsymbol{V}$ denote the query, key, and value matrices 
for this attention head, and $\boldsymbol{M}$ is the causal mask used in decoder-only Transformers. 
When $\gamma=0$, the output of that head is completely ablated, while $\gamma > 1$ amplifies the behavior of the original model. We also perform a renormalization method to keep the output norm stable and prevent generation collapse, following~\cite{zhang2024comprehensive}.

\begin{wrapfigure}{r}{0.35\textwidth}
    \centering
    \small
    \vspace{0.0cm}
    \includegraphics[width=0.35\textwidth]{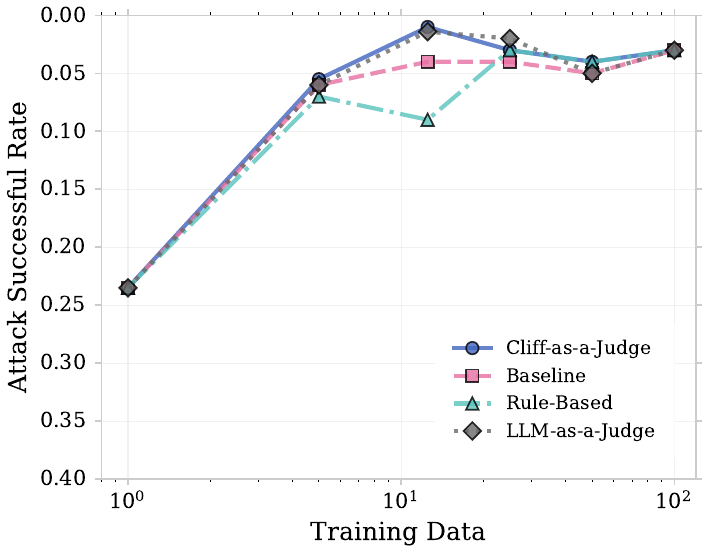}
    \caption{The pareto front beteen Examples and ASR.}
    \vspace{0.3cm}
    \label{tab:pareto}
\end{wrapfigure}

\paragraph{Experiments.} We evaluated our method on JailbreakBench (vanilla attack) and WildJailbreak (adversarial attack) . We test the model performance with ablation on two level: \textit{(i)} \textit{Representation-level:} The refusal score of the prober after the ablation at the last token position on JailbreakBench. \textit{(ii)} \textit{Output-level:} The final Attack Successful Rate after the generation. We defined three thresholds, 1\%, 3\% and 10\%~\footnote{We use 1\% and 3\% for generation (and 3\% and 10\% for refusal score) because ablating a large number of heads may lead to generation collapse.}, as criteria for identifying Refusal Suppression Heads, and set their contributions to zero using the scaling method described in Equation~\ref{eq:zero}. Figure~\ref{fig:trace} presents the ablation results for these Refusal Suppression Heads. Our findings reveal that ablating as few as 10\% of the identified attention heads can more than double the refusal score, while ablating only 3\% of them is sufficient to reduce the probability of producing harmful outputs to below 10\%.

\begin{figure}[t]
    \centering
    \begin{minipage}[t]{0.30\textwidth}
        \centering
        \includegraphics[width=1\linewidth]{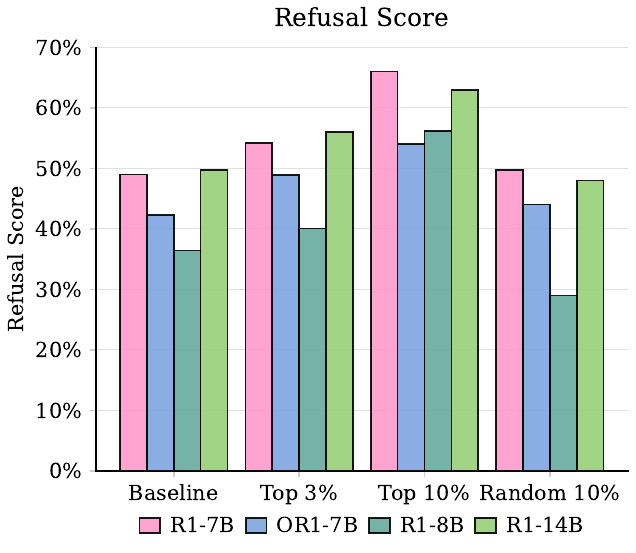}
    \end{minipage}
    \begin{minipage}[t]{0.30\textwidth}
        \centering
        \includegraphics[width=1\linewidth]{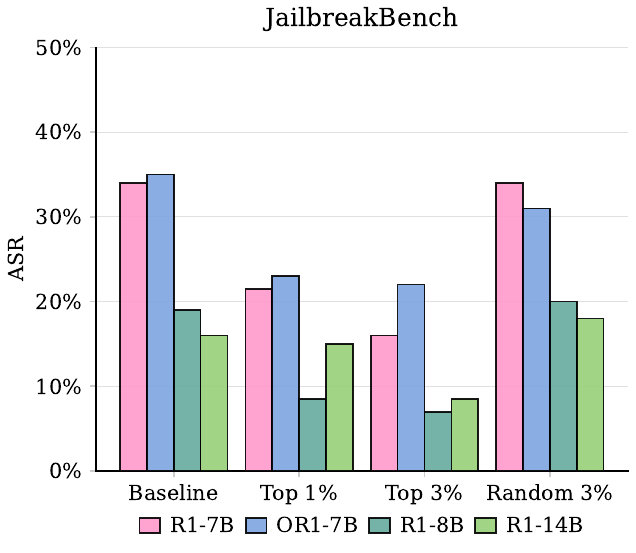}
    \end{minipage}
    \begin{minipage}[t]{0.30\textwidth}
        \centering
        \includegraphics[width=1\linewidth]{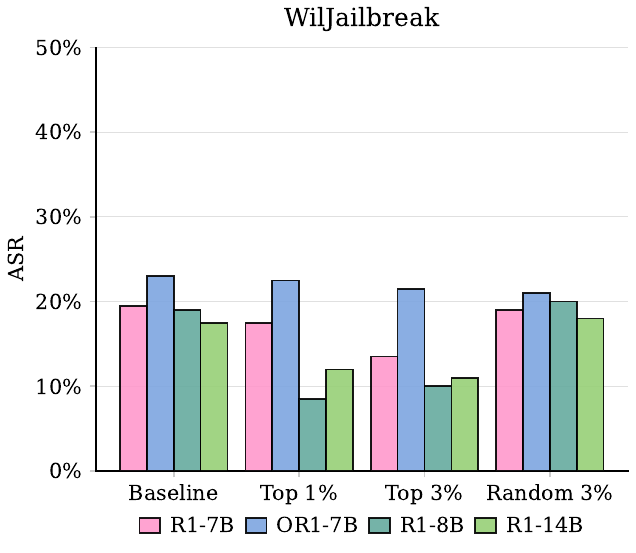}
    \end{minipage}
    \caption{Results of ablating Refusal Suppression Heads. \textit{Left:} The refusal score of prober output \ie \textit{the higher the better}. \textit{Center:} The Attack Successful Rate (ASR) of JailbreakBench \ie \textit{the lower the better.} \textit{Right:} The ASR of WildJailbreak. }
    \label{fig:ablate}
    \vspace{-0.7cm}
\end{figure}

\paragraph{Limitations of Ablation.} We have proposed a seemingly practical solution for tackling the refusal cliff in reasoning models. However, we acknowledge that some readers may remain unconvinced by our conclusions—and rightly so. Intervention-based approaches are not perfect and suffer from several drawbacks:
\textit{(i)} The \textit{superposition} of language model activations~\citep{gao2024scaling} \ie a single activation vector can be expressed as a linear combination of multiple sub-directions corresponding to different task domains, makes it difficult to intervene safely without compromising performance in other domains.
\textit{(ii)} Language models are capable of \textit{self-repair}~\citep{rushing2024explorations}, which further limits the effectiveness of ablation alone in achieving optimal results.
\textit{(iii)} Intervening in a model’s internal components requires redesigned infrastructure and cannot be readily applied to existing tools. We will present a more practical approach in the next section.

\section{Cliff-as-a-Judge: Efficient Alignment via Data Selection}
\label{sec:solutions}

\setlength{\tabcolsep}{1pt}
\setlength{\columnsep}{9pt} 
\setlength{\intextsep}{0pt}

\subsection{Methodology}

\paragraph{Motivations.} From our previous experiments, it is evident that a misaligned reasoning model is not inherently incapable of safe behavior. On the contrary, such a model can often correctly identify the harmful nature of a prompt and internally reflect an intention to refuse during its reasoning process. Under this hypothesis, it follows that aligning an unsafe reasoning model may require only a small set of high-quality alignment examples, thereby achieving a \textit{less-is-more} effect~\citep{zhou2023lima}. 

\begin{wraptable}{r}{0.58\textwidth}
    \centering
    \small
    \vspace{-0.6cm}
    \caption{Comparison between data selection methods.}
    \begin{tabular}{lccc}
    \toprule
         Method&  Continuous&  Judge Model&  Performance\\
         \midrule
         Baseline&  N/A&  None&  Good\\
         Rule-Based& False& None& Moderate\\
         LLM-as-a-judge& False& LLM ($>$1B)& Good\\
         \textbf{Cliff-as-a-judge}& True& Prober ($<$1M)& Good\\
    \bottomrule
    \end{tabular}
    \vspace{0.1cm}
    \label{tab:comp}
\end{wraptable}

\paragraph{Cliff-as-a-judge.} We propose a cliff-based data selection method. Formally, given a dataset \( D \) and a budget \( k \), data selection is to get an optimal subset \( S \subset D, \ |S| = k \) to optimize its alignment performance. Specifically, suppose that, for a given sample, the model’s maximum refusal intention \ie the plateau, expressed within its internal thinking corresponds to a probed refusal score \( I \), and its final generated refusal score is \( I' \) after any cliff drop or suppression. We define the misalignment score $\mathrm{MS} = I - I'$ as a measure of how much the refusal intention expressed in internal reasoning is suppressed in the final output. Intuitively, the most effective subset of alignment data consists of samples with the highest misalignment scores, where training on this data can most efficiently repair safety alignment. Therefore, the optimal selection via Cliff-as-a-judge is:  
\begin{equation}
    \theta^* = \arg\min_{\theta} \ \mathcal{L}_{\text{align}}\!\left(
        \arg\max_{S \subset D,\, |S| = k} \ \frac{1}{k} \sum_{x \in S} \mathrm{MS}(x) \ ; \ \theta
    \right)
\end{equation}
where \(\mathcal{L}_{\text{align}}\) denotes an alignment-oriented objective (\eg Attack Successful Rate). We compare our method with other baselines in Table~\ref{tab:comp}, where Cliff-as-a-judge provides a continuous metric, allows flexible selection of the number of examples, employs a lightweight judge model, and achieves strong performance.

\setlength{\tabcolsep}{3pt}
\begin{table}[t]
    \centering
    \small
    \caption{Benchmark results on safety-related tasks and reasoning-related tasks.}
    \begin{tabular}{llccccccll}
    \toprule
          &&\multicolumn{2}{c}{JailbreakBench}&  \multicolumn{2}{c}{WildJailbreak} & \multicolumn{2}{c}{MMLU-Pro} & \multicolumn{2}{c}{ARC-C}\\
 & & \multicolumn{4}{c}{\textit{Metric: ASR}($\downarrow$
)}& \multicolumn{4}{c}{\textit{Metric: Acc}($\uparrow$
)}\\
         Method &Examples&R1-8B& R1-7B& R1-8B& R1-7B & R1-8B&R1-7B  & R1-8B&R1-7B  \\
        \midrule
        
         Baseline &-&32.0\%& 31.2\%& 19.0\%&  38.0\%& 42.7\%&45.4\%& 40.7\%&41.8\%\\
        Full Training Dataset &$40$k&2.5\%&  1.0\%&  2.0\%&   1.0\%& 40.7\%&42.9\%& 41.3\%&39.9\%\\
         Rule-Based&$21$k$^{\color{Magenta} -46.1\%}$& 1.0\%& 2.5\%& 5.2\%& 2.4\%& 40.8\%&43.4\%& 40.9\%&40.8\%\\
         LLM-as-a-Judge&$5.6$k$^{\color{Magenta}-86.0\%}$&$4.0\%$& 1.5\%& 6.5\%&  1.8\%& 40.8\%&43.7\%& 40.5\%&40.4\%\\
         \rowcolor{orange!10}
         \textbf{Cliff-as-a-judge} &$700^{\color{Magenta}-98.3\%}$& 5.0\%& 3.0\%& 6.0\%& 6.0\%& 41.7\%&44.7\%& 41.4\%&41.2\%\\
    \bottomrule
    \end{tabular}
    \vspace{-0.5cm}
    \label{tab:placeholder}
\end{table}

\subsection{Experiments}

\paragraph{Baselines.} We adopt the training set from WildJailbreak~\citep{wildteaming2024} as our safety alignment corpus with 40k examples. This dataset contains both standard (vanilla) jailbreak attacks and more challenging adversarial jailbreak cases. For baseline data selection methods, we consider: \textit{(i)} full-example training (\ie the unfiltered baseline), \textit{(ii)} rule-based selection~\citep{liu2025roleattentionheads,lab2025safework}, where unsafe cases are identified using keyword matching, \textit{(iii)} LLM-as-a-judge~\citep{gu2024survey,lambert2024tulu,zhang2025realsafe}, where using LlamaGuard~\citep{grattafiori2024llama}. We also select MMLU-Pro~\citep{wang2024mmlu} and ARC-Challenge~\citep{clark2018think} to benchmark the reasoning ability after alignment.

\paragraph{Experimental Results.}
We perform safety fine-tuning on our selected datasets. Table~\ref{tab:placeholder} demonstrates the effectiveness of our Cliff-as-a-judge data selection method across three safety benchmarks. While the baseline models exhibit concerning vulnerabilities with ASR of 19.0-38.0\%, training on the full dataset reduces ASR to 1.0-2.5\%. Remarkably, our method achieves comparable safety performance using only 700 examples (98.3\% data reduction). This substantially outperforms other filtering approaches: Rule-based selection requires 21,566 examples (-46.1\%) and LLM-as-a-judge needs 5,616 examples (-86.0\%) to achieve similar results. As shown in Figure~\ref{tab:pareto}'s Pareto frontier analysis, our approach optimally balances data efficiency with safety performance, translating to reduction in training time while maintaining effective safety alignment across different model architectures. Also, our benchmarking on MMLU-Pro and ARC-C demonstrates that Cliff-as-a-judge is most effective in preserving the model’s original reasoning capabilities, while requiring fewer yet higher-quality examples.

\section{Related Works}

\paragraph{Safety of Large Reasoning Model.}
The development of reasoning models extends safety beyond direct harmfulness classification to deliberate, step-by-step judgment~\citep{wang2025safety} with robustness to jailbreak attempts~\citep{zaremba2025trading, kim2025reasoning}. However, studies also show that this generalization is fragile and can be exploited~\citep{kuo2025h, yan2025thinking, zheng2025beyond, jiang2025safechain}. In response, recent work proposes frameworks—both by evaluating and mitigating risks within reasoning traces themselves~\citep{li2025reasoningshield, zheng2025beyond} and by improving safer outputs~\citep{zhu2025reasoning, jiang2025safechain}. From a different angle, we investigate the mechanistic roots of LRMs’ safety vulnerabilities and offer insights for future solutions.

\paragraph{Mechanistic Interpretability for LLM Safety.}
Mechanistic Interpretability (MI) seeks to reverse-engineer specific model behaviors and functions so their internal mechanisms become human-understandable. Research in this area spans multiple granularities: individual neurons~\citep{gurnee2023finding, stolfo2024confidence}, representations~\citep{marks2024geometry,gurnee2024language}, and larger functional units like MLP~\citep{geva2021Transformer, geva2022Transformer} and attention heads~\citep{mcdougall2023copy, gould2024successor}. Building on these foundations, MI has been increasingly applied to LLM safety~\citep{bereska2024mechanistic}. One thread focuses on representation-level analyses of safety behavior and on techniques for editing safety-related representations~\citep{leong2023self, zou2023representation, arditi2024refusal, cao2025scans, lee2025programming, li2025rethinking, shen2025jailbreak, xu2024uncovering,lee2025xjailbreak}. Another examines components directly implicated in safety, including neurons~\citep{zhao2025understanding}, attention heads~\citep{zhu2024locking, zhou2025on}, and MLPs~\citep{lee2024mechanistic, luo2024jailbreak}. Complementary work studies safety-relevant parameters themselves~\citep{wei2024assessing, yi2025nlsr, gu2025improve}. A parallel line of progress decomposes representations into interpretable, sparse features, enabling automated explanations of safety mechanisms~\citep{minder2025robustly}. These methods suggest promising avenues for achieving more robust safety alignment at the representation level~\citep{liu2024aligning, zou2024improving, rosati2024representation, yin2025constrain}.

\section{Limitations}
While our study sheds light on the mechanistic roots and offers mitigation strategies, several limitations remain. First, our mechanistic analysis focuses primarily on attention heads, leaving other architectural components such as MLP blocks, positional encodings, and cross-layer interactions underexplored. Second, our data‑recipe method depends on having access to the model’s internal representations and refusal scores, which is feasible for open models but may be impractical for proprietary systems. Investigation of proxy metrics or black‑box analogues remains future work.

\section{Conclusions}
In this work, we identified and mechanistically characterized a novel safety failure in large reasoning models -- the \textbf{\term{}}. Through causal tracing, we discovered a small set of Refusal Suppression Heads whose negative contributions are responsible for this phenomenon. Targeted ablation of these heads significantly improves refusal rates, confirming their causal role. Building on these findings, we proposed a targeted safety fine-tuning data recipe that selects training examples most susceptible to the \term. Our experiments show that these methods can improve safety alignment with minimal performance trade‑offs while reducing the training cost. 

\section*{Ethics Statement}
Our research aims to enhance the safety and reliability of Large Reasoning Models (LRMs) by identifying and mitigating a critical failure mode, the ``Refusal Cliff.'' We believe this work contributes positively to the responsible development of AI. However, we acknowledge several ethical considerations inherent in this line of research. Our work involves the analysis of model vulnerabilities to harmful and malicious prompts, which carries a potential dual-use risk. To mitigate this, we have focused on revealing the underlying \textit{mechanisms} of failure rather than developing novel, easily replicable jailbreak techniques. Our proposed solution, ``Cliff-as-a-Judge,'' is a defensive data selection strategy designed to strengthen model safety. The datasets used, such as AdvBench and WildJailbreak, are established benchmarks and were used strictly for evaluating and improving model refusal capabilities without generating new harmful content. We believe our findings can help improve the alignment of models to reduce harmful or biased outputs and encourage the community to build upon our mechanistic insights to develop more robust and ethically aligned AI systems. All research was conducted in adherence to the ICLR Code of Ethics.

\section*{Reproducibility Statement}
We are committed to ensuring the reproducibility of our research. All models used in our experiments (e.g., from the Qwen, DeepSeek, Skywork, and Phi families) and datasets (e.g., AdvBench, JailbreakBench, and WildJailbreak) are publicly available and detailed in Section 2. The implementation details for our core methodologies are provided in the appendix. Specifically, Appendix~\ref{sec:prober_additional} contains the complete setup for training the refusal prober, including hyperparameters and data preprocessing. The procedures for attention head tracing (Section~\ref{sec:head}), head ablation (Section~\ref{sec:head}), and the fine-tuning process for our "Cliff-as-a-Judge" method (Section~\ref{sec:solutions}) are described with sufficient detail for replication. To further facilitate reproducibility, we will release our source code, which includes scripts for data processing, prober training, causal analysis, and model fine-tuning, upon publication of this paper.


\bibliography{iclr2026_conference}
\bibliographystyle{iclr2026_conference}
\appendix

\section*{Use of LLMs} In the preparation of this manuscript, we utilized LLMs as a writing assistant. The use of LLMs was confined to tasks such as improving grammar, refining phrasing for clarity, and polishing the overall language of the paper. All core scientific contributions, including the conceptualization of ideas, the design and execution of experiments, the analysis of results, and the conclusions, are entirely the work of the human authors. The authors bear full responsibility for the content and claims presented in this work.

\section{Prober}
\label{sec:prober_additional}

In this section, we provide a detailed description of the architecture, data collection, and training procedure for the refusal prober used in our experiments. This prober is a linear classifier designed to predict whether a model will refuse a harmful request based on its internal hidden states.

\paragraph{Prober Architecture.}
The prober is implemented as a simple linear classifier. Given a hidden state vector $\vh \in \mathbb{R}^d$ from the reasoning model, where $d$ is the hidden dimension size, the prober computes a single logit. This is followed by a sigmoid function to produce a refusal probability, as defined in Equation~\ref{eq:prober}. The model is implemented in PyTorch using a single `torch.nn.Linear` layer. We use the Binary Cross-Entropy with Logits loss function (`nn.BCEWithLogitsLoss`) for training, which is numerically stable and suitable for binary classification tasks.

\paragraph{Dataset Collection and Preprocessing.}
To train the prober, we constructed a balanced dataset of hidden states corresponding to both refusal and non-refusal responses.
\begin{itemize}[leftmargin=0.6cm, itemsep=0.05cm]
    \item \textbf{Refusal Examples (Positive Class):} We collected examples where the model refused to comply with a harmful prompt. These were sourced from the AdvBench dataset~\citep{zou2023universal}. An output was labeled as a refusal if it contained keywords like ``I'm sorry,'' ``I cannot,'' or similar phrases within the first 32 tokens of the response.
    \item \textbf{Non-Refusal Examples (Negative Class):} For the non-refusal class, we used harmless prompts and their corresponding compliant answers from the UltraChat-SFT dataset~\citep{ding2023enhancing}.
\end{itemize}
For each example in both classes, we fed the full input sequence (user prompt + model's chain of thought + thinking-end template) into the target reasoning model. We then extracted the hidden state vector from the **final token position** at the **last transformer layer**. These hidden state vectors form the training data for our prober.

\paragraph{Training Details.}
The prober was trained on the collected hidden states. Before training, we balanced the dataset by randomly downsampling the larger class to match the number of samples in the smaller class, ensuring an equal number of refusal and non-refusal examples. The full dataset was then split into training (80\%) and validation (20\%) sets.

The training hyperparameters are as follows:
\begin{itemize}[leftmargin=0.6cm, itemsep=0.05cm]
    \item \textbf{Optimizer:} Adam
    \item \textbf{Learning Rate:} $1 \times 10^{-3}$
    \item \textbf{Batch Size:} 256
    \item \textbf{Epochs:} 5
\end{itemize}
We selected the model checkpoint that achieved the highest accuracy on the validation set. As reported in Section~\ref{sec:cliff}, the final prober achieved over 95\% validation accuracy on in-distribution data and demonstrated strong generalization to an out-of-distribution (OOD) dataset, JailbreakBench. This high accuracy confirms that the prober reliably captures the model's refusal intention from its final hidden state.

\section{Supervised Fine-tuning Details}
\label{sec:appendix_training_details}

We performed full-parameter supervised fine-tuning (SFT) to repair the safety alignment of the reasoning models using the data subsets selected by our Cliff-as-a-Judge method. The entire training process was conducted using the LLaMA-Factory library. The base model for the fine-tuning experiments reported in Section~\ref{sec:solutions} was \texttt{deepseek-ai/DeepSeek-R1-Distill-Qwen-7B}. We utilized DeepSpeed ZeRO Stage 2 for efficient distributed training. The key hyperparameters and configuration settings are detailed below:

\begin{itemize}[leftmargin=*, itemsep=0.05cm]
    \item \textbf{Finetuning Type:} Full-parameter SFT
    \item \textbf{Learning Rate:} $5 \times 10^{-6}$
    \item \textbf{LR Scheduler:} Linear
    \item \textbf{Epochs:} 1.0
    \item \textbf{Batch Size:} 1 per device with 4 gradient accumulation steps, resulting in an effective batch size of 4.
    \item \textbf{Optimizer:} AdamW (\texttt{adamw\_torch})
    \item \textbf{Precision:} BF16
    \item \textbf{Max Sequence Length:} 16,384
    \item \textbf{Attention Implementation:} Flash Attention
    \item \textbf{Prompt Template:} \texttt{deepseekr1}
    \item \textbf{Distributed Training:} DeepSpeed ZeRO Stage 2
\end{itemize}

\end{document}

%% file: math_commands.tex
\usepackage{amsmath,amsfonts,bm}

\def\eqref#1{equation~\ref{#1}}

\def\1{\bm{1}}

\def\vh{{\bm{h}}}

\DeclareMathAlphabet{\mathsfit}{\encodingdefault}{\sfdefault}{m}{sl}
\SetMathAlphabet{\mathsfit}{bold}{\encodingdefault}{\sfdefault}{bx}{n}













%% file: iclr2026_conference.bbl
\begin{thebibliography}{87}
\providecommand{\natexlab}[1]{#1}
\providecommand{\url}[1]{\texttt{#1}}
\expandafter\ifx\csname urlstyle\endcsname\relax
  \providecommand{\doi}[1]{doi: #1}\else
  \providecommand{\doi}{doi: \begingroup \urlstyle{rm}\Url}\fi

\bibitem[Abdin et~al.(2024)Abdin, Aneja, Behl, Bubeck, Eldan, Gunasekar, Harrison, Hewett, Javaheripi, Kauffmann, et~al.]{abdin2024phi}
Marah Abdin, Jyoti Aneja, Harkirat Behl, S{\'e}bastien Bubeck, Ronen Eldan, Suriya Gunasekar, Michael Harrison, Russell~J Hewett, Mojan Javaheripi, Piero Kauffmann, et~al.
\newblock Phi-4 technical report.
\newblock \emph{arXiv preprint arXiv:2412.08905}, 2024.

\bibitem[Abdin et~al.(2025)Abdin, Agarwal, Awadallah, Balachandran, Behl, Chen, de~Rosa, Gunasekar, Javaheripi, Joshi, et~al.]{abdin2025phi}
Marah Abdin, Sahaj Agarwal, Ahmed Awadallah, Vidhisha Balachandran, Harkirat Behl, Lingjiao Chen, Gustavo de~Rosa, Suriya Gunasekar, Mojan Javaheripi, Neel Joshi, et~al.
\newblock Phi-4-reasoning technical report.
\newblock \emph{arXiv preprint arXiv:2504.21318}, 2025.

\bibitem[Allan(2018)]{allan2018hermes}
Arlene Allan.
\newblock \emph{Hermes}.
\newblock Routledge, 2018.

\bibitem[Arcuschin et~al.(2025)Arcuschin, Janiak, Krzyzanowski, Rajamanoharan, Nanda, and Conmy]{Arcuschin2025ChainofThoughtRI}
Iv'an Arcuschin, Jett Janiak, Robert Krzyzanowski, Senthooran Rajamanoharan, Neel Nanda, and Arthur Conmy.
\newblock Chain-of-thought reasoning in the wild is not always faithful.
\newblock In \emph{Reasoning and Planning for LLMs @ ICLR2025}, 2025.

\bibitem[Arditi et~al.(2024)Arditi, Obeso, Syed, Paleka, Panickssery, Gurnee, and Nanda]{arditi2024refusal}
Andy Arditi, Oscar Obeso, Aaquib Syed, Daniel Paleka, Nina Panickssery, Wes Gurnee, and Neel Nanda.
\newblock Refusal in language models is mediated by a single direction.
\newblock \emph{Advances in Neural Information Processing Systems}, 37:\penalty0 136037--136083, 2024.

\bibitem[Barez et~al.(2025)Barez, Wu, Arcuschin, Lan, Wang, Siegel, Collignon, Neo, Lee, Paren, Bibi, Trager, Fornasiere, Yan, Elazar, and Bengio]{barez-chain-2025}
Fazl Barez, Tung-Yu Wu, Iván Arcuschin, Michael Lan, Vincent Wang, Noah Siegel, Nicolas Collignon, Clement Neo, Isabelle Lee, Alasdair Paren, Adel Bibi, Robert Trager, Damiano Fornasiere, John Yan, Yanai Elazar, and Yoshua Bengio.
\newblock Chain-of-thought is not explainability, 2025.

\bibitem[Bereska \& Gavves(2024)Bereska and Gavves]{bereska2024mechanistic}
Leonard Bereska and Stratis Gavves.
\newblock Mechanistic interpretability for {AI} safety - a review.
\newblock \emph{Transactions on Machine Learning Research}, 2024.
\newblock ISSN 2835-8856.
\newblock URL \url{https://openreview.net/forum?id=ePUVetPKu6}.
\newblock Survey Certification, Expert Certification.

\bibitem[Cao et~al.(2025)Cao, Yang, and Zhao]{cao2025scans}
Zouying Cao, Yifei Yang, and Hai Zhao.
\newblock Scans: Mitigating the exaggerated safety for llms via safety-conscious activation steering.
\newblock In \emph{Proceedings of the AAAI Conference on Artificial Intelligence}, volume~39, pp.\  23523--23531, 2025.

\bibitem[Chan et~al.(2025)Chan, Yong, and Bach]{chan2025can}
Yik~Siu Chan, Zheng-Xin Yong, and Stephen~H Bach.
\newblock Can we predict alignment before models finish thinking? towards monitoring misaligned reasoning models.
\newblock \emph{arXiv preprint arXiv:2507.12428}, 2025.

\bibitem[Chao et~al.(2024)Chao, Debenedetti, Robey, Andriushchenko, Croce, Sehwag, Dobriban, Flammarion, Pappas, Tramer, et~al.]{chao2024jailbreakbench}
Patrick Chao, Edoardo Debenedetti, Alexander Robey, Maksym Andriushchenko, Francesco Croce, Vikash Sehwag, Edgar Dobriban, Nicolas Flammarion, George~J Pappas, Florian Tramer, et~al.
\newblock Jailbreakbench: An open robustness benchmark for jailbreaking large language models.
\newblock \emph{Advances in Neural Information Processing Systems}, 37:\penalty0 55005--55029, 2024.

\bibitem[Clark et~al.(2018)Clark, Cowhey, Etzioni, Khot, Sabharwal, Schoenick, and Tafjord]{clark2018think}
Peter Clark, Isaac Cowhey, Oren Etzioni, Tushar Khot, Ashish Sabharwal, Carissa Schoenick, and Oyvind Tafjord.
\newblock Think you have solved question answering? try arc, the ai2 reasoning challenge.
\newblock \emph{arXiv preprint arXiv:1803.05457}, 2018.

\bibitem[Ding et~al.(2023)Ding, Chen, Xu, Qin, Hu, Liu, Sun, and Zhou]{ding2023enhancing}
Ning Ding, Yulin Chen, Bokai Xu, Yujia Qin, Shengding Hu, Zhiyuan Liu, Maosong Sun, and Bowen Zhou.
\newblock Enhancing chat language models by scaling high-quality instructional conversations.
\newblock In \emph{The 2023 Conference on Empirical Methods in Natural Language Processing}, 2023.
\newblock URL \url{https://openreview.net/forum?id=oEsYs3WRc3}.

\bibitem[Engels et~al.(2025)Engels, Michaud, Liao, Gurnee, and Tegmark]{engels2025not}
Joshua Engels, Eric~J Michaud, Isaac Liao, Wes Gurnee, and Max Tegmark.
\newblock Not all language model features are one-dimensionally linear.
\newblock In \emph{The Thirteenth International Conference on Learning Representations}, 2025.
\newblock URL \url{https://openreview.net/forum?id=d63a4AM4hb}.

\bibitem[Gao et~al.(2024)Gao, la~Tour, Tillman, Goh, Troll, Radford, Sutskever, Leike, and Wu]{gao2024scaling}
Leo Gao, Tom~Dupr{\'e} la~Tour, Henk Tillman, Gabriel Goh, Rajan Troll, Alec Radford, Ilya Sutskever, Jan Leike, and Jeffrey Wu.
\newblock Scaling and evaluating sparse autoencoders.
\newblock \emph{arXiv preprint arXiv:2406.04093}, 2024.

\bibitem[Geva et~al.(2021)Geva, Schuster, Berant, and Levy]{geva2021Transformer}
Mor Geva, Roei Schuster, Jonathan Berant, and Omer Levy.
\newblock Transformer feed-forward layers are key-value memories.
\newblock In Marie-Francine Moens, Xuanjing Huang, Lucia Specia, and Scott Wen-tau Yih (eds.), \emph{Proceedings of the 2021 Conference on Empirical Methods in Natural Language Processing}, pp.\  5484--5495, Online and Punta Cana, Dominican Republic, November 2021. Association for Computational Linguistics.
\newblock \doi{10.18653/v1/2021.emnlp-main.446}.
\newblock URL \url{https://aclanthology.org/2021.emnlp-main.446/}.

\bibitem[Geva et~al.(2022)Geva, Caciularu, Wang, and Goldberg]{geva2022Transformer}
Mor Geva, Avi Caciularu, Kevin Wang, and Yoav Goldberg.
\newblock Transformer feed-forward layers build predictions by promoting concepts in the vocabulary space.
\newblock In Yoav Goldberg, Zornitsa Kozareva, and Yue Zhang (eds.), \emph{Proceedings of the 2022 Conference on Empirical Methods in Natural Language Processing}, pp.\  30--45, Abu Dhabi, United Arab Emirates, December 2022. Association for Computational Linguistics.
\newblock \doi{10.18653/v1/2022.emnlp-main.3}.
\newblock URL \url{https://aclanthology.org/2022.emnlp-main.3/}.

\bibitem[Gorton \& Lewis(2025)Gorton and Lewis]{gorton2025adversarialexamplesnot}
Liv Gorton and Owen Lewis.
\newblock Adversarial examples are not bugs, they are superposition, 2025.
\newblock URL \url{https://arxiv.org/abs/2508.17456}.

\bibitem[Gould et~al.(2024)Gould, Ong, Ogden, and Conmy]{gould2024successor}
Rhys Gould, Euan Ong, George Ogden, and Arthur Conmy.
\newblock Successor heads: Recurring, interpretable attention heads in the wild.
\newblock In \emph{The Twelfth International Conference on Learning Representations}, 2024.
\newblock URL \url{https://openreview.net/forum?id=kvcbV8KQsi}.

\bibitem[Grattafiori et~al.(2024)Grattafiori, Dubey, Jauhri, Pandey, Kadian, Al-Dahle, Letman, Mathur, Schelten, Vaughan, et~al.]{grattafiori2024llama}
Aaron Grattafiori, Abhimanyu Dubey, Abhinav Jauhri, Abhinav Pandey, Abhishek Kadian, Ahmad Al-Dahle, Aiesha Letman, Akhil Mathur, Alan Schelten, Alex Vaughan, et~al.
\newblock The llama 3 herd of models.
\newblock \emph{arXiv preprint arXiv:2407.21783}, 2024.

\bibitem[Gu et~al.(2024)Gu, Jiang, Shi, Tan, Zhai, Xu, Li, Shen, Ma, Liu, et~al.]{gu2024survey}
Jiawei Gu, Xuhui Jiang, Zhichao Shi, Hexiang Tan, Xuehao Zhai, Chengjin Xu, Wei Li, Yinghan Shen, Shengjie Ma, Honghao Liu, et~al.
\newblock A survey on llm-as-a-judge.
\newblock \emph{arXiv preprint arXiv:2411.15594}, 2024.

\bibitem[Gu et~al.(2025)Gu, Wang, and Mao]{gu2025improve}
Peijian Gu, Quan Wang, and Zhendong Mao.
\newblock Improve safety training of large language models with safety-critical singular vectors localization.
\newblock In Wanxiang Che, Joyce Nabende, Ekaterina Shutova, and Mohammad~Taher Pilehvar (eds.), \emph{Proceedings of the 63rd Annual Meeting of the Association for Computational Linguistics (Volume 1: Long Papers)}, pp.\  4941--4954, Vienna, Austria, July 2025. Association for Computational Linguistics.
\newblock ISBN 979-8-89176-251-0.
\newblock \doi{10.18653/v1/2025.acl-long.245}.
\newblock URL \url{https://aclanthology.org/2025.acl-long.245/}.

\bibitem[Guo et~al.(2025)Guo, Yang, Zhang, Song, Zhang, Xu, Zhu, Ma, Wang, Bi, et~al.]{guo2025deepseek}
Daya Guo, Dejian Yang, Haowei Zhang, Junxiao Song, Ruoyu Zhang, Runxin Xu, Qihao Zhu, Shirong Ma, Peiyi Wang, Xiao Bi, et~al.
\newblock Deepseek-r1: Incentivizing reasoning capability in llms via reinforcement learning.
\newblock \emph{arXiv preprint arXiv:2501.12948}, 2025.

\bibitem[Gurnee \& Tegmark(2024)Gurnee and Tegmark]{gurnee2024language}
Wes Gurnee and Max Tegmark.
\newblock Language models represent space and time.
\newblock In \emph{The Twelfth International Conference on Learning Representations}, 2024.
\newblock URL \url{https://openreview.net/forum?id=jE8xbmvFin}.

\bibitem[Gurnee et~al.(2023)Gurnee, Nanda, Pauly, Harvey, Troitskii, and Bertsimas]{gurnee2023finding}
Wes Gurnee, Neel Nanda, Matthew Pauly, Katherine Harvey, Dmitrii Troitskii, and Dimitris Bertsimas.
\newblock Finding neurons in a haystack: Case studies with sparse probing.
\newblock \emph{Transactions on Machine Learning Research}, 2023.
\newblock ISSN 2835-8856.
\newblock URL \url{https://openreview.net/forum?id=JYs1R9IMJr}.

\bibitem[He et~al.(2025)He, Liu, Liu, Yan, Wang, Cheng, Zhang, Zhang, Xu, Shen, et~al.]{he2025skywork}
Jujie He, Jiacai Liu, Chris~Yuhao Liu, Rui Yan, Chaojie Wang, Peng Cheng, Xiaoyu Zhang, Fuxiang Zhang, Jiacheng Xu, Wei Shen, et~al.
\newblock Skywork open reasoner 1 technical report.
\newblock \emph{arXiv preprint arXiv:2505.22312}, 2025.

\bibitem[Heimersheim \& Nanda(2024)Heimersheim and Nanda]{heimersheim2024use}
Stefan Heimersheim and Neel Nanda.
\newblock How to use and interpret activation patching.
\newblock \emph{arXiv preprint arXiv:2404.15255}, 2024.

\bibitem[Hendel et~al.(2023)Hendel, Geva, and Globerson]{Hendel2023InContextLC}
Roee Hendel, Mor Geva, and Amir Globerson.
\newblock In-context learning creates task vectors.
\newblock \emph{ArXiv}, abs/2310.15916, 2023.

\bibitem[{Hugging Face}(2025)]{openr1}
{Hugging Face}.
\newblock Open r1: A fully open reproduction of deepseek-r1, January 2025.
\newblock URL \url{https://github.com/huggingface/open-r1}.

\bibitem[Ilharco et~al.(2022)Ilharco, Ribeiro, Wortsman, Gururangan, Schmidt, Hajishirzi, and Farhadi]{Ilharco2022EditingMW}
Gabriel Ilharco, Marco~Tulio Ribeiro, Mitchell Wortsman, Suchin Gururangan, Ludwig Schmidt, Hannaneh Hajishirzi, and Ali Farhadi.
\newblock Editing models with task arithmetic.
\newblock \emph{ArXiv}, abs/2212.04089, 2022.

\bibitem[Jiang et~al.(2025)Jiang, Xu, Li, Niu, Xiang, Li, Lin, and Poovendran]{jiang2025safechain}
Fengqing Jiang, Zhangchen Xu, Yuetai Li, Luyao Niu, Zhen Xiang, Bo~Li, Bill~Yuchen Lin, and Radha Poovendran.
\newblock Safechain: Safety of language models with long chain-of-thought reasoning capabilities.
\newblock \emph{arXiv preprint arXiv:2502.12025}, 2025.

\bibitem[Jiang et~al.(2024)Jiang, Rao, Han, Ettinger, Brahman, Kumar, Mireshghallah, Lu, Sap, Choi, and Dziri]{wildteaming2024}
Liwei Jiang, Kavel Rao, Seungju Han, Allyson Ettinger, Faeze Brahman, Sachin Kumar, Niloofar Mireshghallah, Ximing Lu, Maarten Sap, Yejin Choi, and Nouha Dziri.
\newblock Wildteaming at scale: From in-the-wild jailbreaks to (adversarially) safer language models, 2024.
\newblock URL \url{https://arxiv.org/abs/2406.18510}.

\bibitem[Kim et~al.(2025)Kim, Tajwar, Raghunathan, and Kumar]{kim2025reasoning}
Taeyoun Kim, Fahim Tajwar, Aditi Raghunathan, and Aviral Kumar.
\newblock Reasoning as an adaptive defense for safety.
\newblock \emph{arXiv preprint arXiv:2507.00971}, 2025.

\bibitem[Kuo et~al.(2025)Kuo, Zhang, Ding, Wang, DiValentin, Bao, Wei, Li, and Chen]{kuo2025h}
Martin Kuo, Jianyi Zhang, Aolin Ding, Qinsi Wang, Louis DiValentin, Yujia Bao, Wei Wei, Hai Li, and Yiran Chen.
\newblock H-cot: Hijacking the chain-of-thought safety reasoning mechanism to jailbreak large reasoning models, including openai o1/o3, deepseek-r1, and gemini 2.0 flash thinking.
\newblock \emph{arXiv preprint arXiv:2502.12893}, 2025.

\bibitem[Lab et~al.(2025)Lab, Bao, Chen, Chen, Chen, Chen, Chen, Chen, Chen, Cheng, et~al.]{lab2025safework}
Shanghai~AI Lab, Yicheng Bao, Guanxu Chen, Mingkang Chen, Yunhao Chen, Chiyu Chen, Lingjie Chen, Sirui Chen, Xinquan Chen, Jie Cheng, et~al.
\newblock Safework-r1: Coevolving safety and intelligence under the ai-45 law.
\newblock \emph{arXiv preprint arXiv:2507.18576}, 2025.

\bibitem[Lambert et~al.(2024)Lambert, Morrison, Pyatkin, Huang, Ivison, Brahman, Miranda, Liu, Dziri, Lyu, et~al.]{lambert2024tulu}
Nathan Lambert, Jacob Morrison, Valentina Pyatkin, Shengyi Huang, Hamish Ivison, Faeze Brahman, Lester James~V Miranda, Alisa Liu, Nouha Dziri, Shane Lyu, et~al.
\newblock Tulu 3: Pushing frontiers in open language model post-training.
\newblock \emph{arXiv preprint arXiv:2411.15124}, 2024.

\bibitem[Lee et~al.(2024)Lee, Bai, Pres, Wattenberg, Kummerfeld, and Mihalcea]{lee2024mechanistic}
Andrew Lee, Xiaoyan Bai, Itamar Pres, Martin Wattenberg, Jonathan~K. Kummerfeld, and Rada Mihalcea.
\newblock A mechanistic understanding of alignment algorithms: A case study on {DPO} and toxicity.
\newblock In \emph{Forty-first International Conference on Machine Learning}, 2024.
\newblock URL \url{https://openreview.net/forum?id=dBqHGZPGZI}.

\bibitem[Lee et~al.(2025{\natexlab{a}})Lee, Padhi, Ramamurthy, Miehling, Dognin, Nagireddy, and Dhurandhar]{lee2025programming}
Bruce~W. Lee, Inkit Padhi, Karthikeyan~Natesan Ramamurthy, Erik Miehling, Pierre Dognin, Manish Nagireddy, and Amit Dhurandhar.
\newblock Programming refusal with conditional activation steering.
\newblock In \emph{The Thirteenth International Conference on Learning Representations}, 2025{\natexlab{a}}.
\newblock URL \url{https://openreview.net/forum?id=Oi47wc10sm}.

\bibitem[Lee et~al.(2025{\natexlab{b}})Lee, Ni, Wei, Li, Fan, Argha, Alinejad-Rokny, Xu, Gong, and Yang]{lee2025xjailbreak}
Sunbowen Lee, Shiwen Ni, Chi Wei, Shuaimin Li, Liyang Fan, Ahmadreza Argha, Hamid Alinejad-Rokny, Ruifeng Xu, Yicheng Gong, and Min Yang.
\newblock xjailbreak: Representation space guided reinforcement learning for interpretable llm jailbreaking.
\newblock \emph{arXiv preprint arXiv:2501.16727}, 2025{\natexlab{b}}.

\bibitem[Leong et~al.(2023)Leong, Cheng, Wang, Wang, and Li]{leong2023self}
Chak~Tou Leong, Yi~Cheng, Jiashuo Wang, Jian Wang, and Wenjie Li.
\newblock Self-detoxifying language models via toxification reversal.
\newblock In Houda Bouamor, Juan Pino, and Kalika Bali (eds.), \emph{Proceedings of the 2023 Conference on Empirical Methods in Natural Language Processing}, pp.\  4433--4449, Singapore, December 2023. Association for Computational Linguistics.
\newblock \doi{10.18653/v1/2023.emnlp-main.269}.
\newblock URL \url{https://aclanthology.org/2023.emnlp-main.269/}.

\bibitem[Li et~al.(2025{\natexlab{a}})Li, Mo, Li, Wang, and Wang]{li2025smarter}
Ang Li, Yichuan Mo, Mingjie Li, Yifei Wang, and Yisen Wang.
\newblock Are smarter llms safer? exploring safety-reasoning trade-offs in prompting and fine-tuning.
\newblock \emph{arXiv preprint arXiv:2502.09673}, 2025{\natexlab{a}}.

\bibitem[Li et~al.(2025{\natexlab{b}})Li, Wang, Pan, Hong, and Yang]{li2025reasoningshield}
Changyi Li, Jiayi Wang, Xudong Pan, Geng Hong, and Min Yang.
\newblock Reasoningshield: Content safety detection over reasoning traces of large reasoning models.
\newblock \emph{arXiv preprint arXiv:2505.17244}, 2025{\natexlab{b}}.

\bibitem[Li et~al.(2025{\natexlab{c}})Li, Wang, Liu, Wu, Dou, Lv, Wang, Zheng, and Huang]{li2025rethinking}
Tianlong Li, Zhenghua Wang, Wenhao Liu, Muling Wu, Shihan Dou, Changze Lv, Xiaohua Wang, Xiaoqing Zheng, and Xuanjing Huang.
\newblock Revisiting jailbreaking for large language models: A representation engineering perspective.
\newblock In Owen Rambow, Leo Wanner, Marianna Apidianaki, Hend Al-Khalifa, Barbara~Di Eugenio, and Steven Schockaert (eds.), \emph{Proceedings of the 31st International Conference on Computational Linguistics}, pp.\  3158--3178, Abu Dhabi, UAE, January 2025{\natexlab{c}}. Association for Computational Linguistics.
\newblock URL \url{https://aclanthology.org/2025.coling-main.212/}.

\bibitem[Liu et~al.(2025{\natexlab{a}})Liu, Wu, Zhou, Yu, Wang, Lin, Zhang, Chen, Zhang, Zhang, Liu, Tong, Wang, Wei, Yan, Song, Ma, Yue, Qiao, Lin, Zhang, Zhu, Sheng, Zhang, Dai, Zhu, Liu, Yuan, Chen, Ma, Zhu, Fan, Zuo, Liu, and Yu]{liu2025dapoopensourcellm}
Jingjing Liu, Yonghui Wu, Hao Zhou, Qiying Yu, Chengyi Wang, Zhiqi Lin, Chi Zhang, Jiangjie Chen, Ya-Qin Zhang, Zheng Zhang, Xin Liu, Yuxuan Tong, Mingxuan Wang, Xiangpeng Wei, Lin Yan, Yuxuan Song, Wei-Ying Ma, Yu~Yue, Mu~Qiao, Haibin Lin, Mofan Zhang, Jinhua Zhu, Guangming Sheng, Wang Zhang, Weinan Dai, Hang Zhu, Gaohong Liu, Yufeng Yuan, Jiaze Chen, Bole Ma, Ruofei Zhu, Tiantian Fan, Xiaochen Zuo, Lingjun Liu, and Hongli Yu.
\newblock Dapo: An open-source llm reinforcement learning system at scale, 2025{\natexlab{a}}.
\newblock URL \url{https://arxiv.org/abs/2503.14476}.

\bibitem[Liu et~al.(2024)Liu, Wang, Wu, Li, Lv, Ling, JianHao, Zhang, Zheng, and Huang]{liu2024aligning}
Wenhao Liu, Xiaohua Wang, Muling Wu, Tianlong Li, Changze Lv, Zixuan Ling, Zhu JianHao, Cenyuan Zhang, Xiaoqing Zheng, and Xuanjing Huang.
\newblock Aligning large language models with human preferences through representation engineering.
\newblock In Lun-Wei Ku, Andre Martins, and Vivek Srikumar (eds.), \emph{Proceedings of the 62nd Annual Meeting of the Association for Computational Linguistics (Volume 1: Long Papers)}, pp.\  10619--10638, Bangkok, Thailand, August 2024. Association for Computational Linguistics.
\newblock \doi{10.18653/v1/2024.acl-long.572}.
\newblock URL \url{https://aclanthology.org/2024.acl-long.572/}.

\bibitem[Liu et~al.(2025{\natexlab{b}})Liu, Yu, Huang, Li, Xu, Wang, Zhou, Zhang, and Fang]{liu2025roleattentionheads}
Yang Liu, Haiyang Yu, Fei Huang, Yongbin Li, Rongwu Xu, Kun Wang, Zhenhong Zhou, Xinghua Zhang, and Junfeng Fang.
\newblock On the role of attention heads in large language model safety, 2025{\natexlab{b}}.
\newblock URL \url{https://arxiv.org/abs/2410.13708}.

\bibitem[Luo et~al.(2024)Luo, Zhou, Wang, and Dong]{luo2024jailbreak}
Yifan Luo, Zhennan Zhou, Meitan Wang, and Bin Dong.
\newblock Jailbreak instruction-tuned llms via end-of-sentence mlp re-weighting.
\newblock \emph{arXiv preprint arXiv:2410.10150}, 2024.

\bibitem[Marks \& Tegmark(2024)Marks and Tegmark]{marks2024geometry}
Samuel Marks and Max Tegmark.
\newblock The geometry of truth: Emergent linear structure in large language model representations of true/false datasets.
\newblock In \emph{First Conference on Language Modeling}, 2024.
\newblock URL \url{https://openreview.net/forum?id=aajyHYjjsk}.

\bibitem[McDougall et~al.(2023)McDougall, Conmy, Rushing, McGrath, and Nanda]{mcdougall2023copy}
Callum McDougall, Arthur Conmy, Cody Rushing, Thomas McGrath, and Neel Nanda.
\newblock Copy suppression: Comprehensively understanding an attention head.
\newblock \emph{arXiv preprint arXiv:2310.04625}, 2023.

\bibitem[Minder et~al.(2025)Minder, Dumas, Chughtai, and Nanda]{minder2025robustly}
Julian Minder, Cl{\'e}ment Dumas, Bilal Chughtai, and Neel Nanda.
\newblock Robustly identifying concepts introduced during chat fine-tuning using crosscoders.
\newblock In \emph{Sparsity in LLMs (SLLM): Deep Dive into Mixture of Experts, Quantization, Hardware, and Inference}, 2025.

\bibitem[Nanda et~al.(2025)Nanda, Tegmark, Rajamanoharan, Engels, and Kantamneni]{nanda2025sparseautoencodersuseful}
Neel Nanda, Max Tegmark, Senthooran Rajamanoharan, Joshua Engels, and Subhash Kantamneni.
\newblock Are sparse autoencoders useful? a case study in sparse probing, 2025.
\newblock URL \url{https://arxiv.org/abs/2502.16681}.

\bibitem[Olsson et~al.(2022)Olsson, Elhage, Nanda, Joseph, DasSarma, Henighan, Mann, Askell, Bai, Chen, et~al.]{olsson2022context}
Catherine Olsson, Nelson Elhage, Neel Nanda, Nicholas Joseph, Nova DasSarma, Tom Henighan, Ben Mann, Amanda Askell, Yuntao Bai, Anna Chen, et~al.
\newblock In-context learning and induction heads.
\newblock \emph{arXiv preprint arXiv:2209.11895}, 2022.

\bibitem[Rosati et~al.(2024)Rosati, Wehner, Williams, Bartoszcze, Gonzales, carsten maple, Majumdar, Sajjad, and Rudzicz]{rosati2024representation}
Domenic Rosati, Jan Wehner, Kai Williams, Lukasz Bartoszcze, Robie Gonzales, carsten maple, Subhabrata Majumdar, Hassan Sajjad, and Frank Rudzicz.
\newblock Representation noising: A defence mechanism against harmful finetuning.
\newblock In \emph{The Thirty-eighth Annual Conference on Neural Information Processing Systems}, 2024.
\newblock URL \url{https://openreview.net/forum?id=eP9auEJqFg}.

\bibitem[Rushing \& Nanda(2024)Rushing and Nanda]{rushing2024explorations}
Cody Rushing and Neel Nanda.
\newblock Explorations of self-repair in language models.
\newblock \emph{arXiv preprint arXiv:2402.15390}, 2024.

\bibitem[Sabbaghi et~al.(2025)Sabbaghi, Kassianik, Pappas, Singer, Karbasi, and Hassani]{sabbaghi2025adversarial}
Mahdi Sabbaghi, Paul Kassianik, George Pappas, Yaron Singer, Amin Karbasi, and Hamed Hassani.
\newblock Adversarial reasoning at jailbreaking time.
\newblock \emph{arXiv preprint arXiv:2502.01633}, 2025.

\bibitem[Shao et~al.(2024)Shao, Wang, Zhu, Xu, Song, Bi, Zhang, Zhang, Li, Wu, et~al.]{shao2024deepseekmath}
Zhihong Shao, Peiyi Wang, Qihao Zhu, Runxin Xu, Junxiao Song, Xiao Bi, Haowei Zhang, Mingchuan Zhang, YK~Li, Yang Wu, et~al.
\newblock Deepseekmath: Pushing the limits of mathematical reasoning in open language models.
\newblock \emph{arXiv preprint arXiv:2402.03300}, 2024.

\bibitem[Shen et~al.(2025)Shen, Zhao, Dong, He, and Zeng]{shen2025jailbreak}
Guobin Shen, Dongcheng Zhao, Yiting Dong, Xiang He, and Yi~Zeng.
\newblock Jailbreak antidote: Runtime safety-utility balance via sparse representation adjustment in large language models.
\newblock In \emph{The Thirteenth International Conference on Learning Representations}, 2025.
\newblock URL \url{https://openreview.net/forum?id=s20W12XTF8}.

\bibitem[Stolfo et~al.(2024)Stolfo, Wu, Gurnee, Belinkov, Song, Sachan, and Nanda]{stolfo2024confidence}
Alessandro Stolfo, Ben~Peng Wu, Wes Gurnee, Yonatan Belinkov, Xingyi Song, Mrinmaya Sachan, and Neel Nanda.
\newblock Confidence regulation neurons in language models.
\newblock In \emph{The Thirty-eighth Annual Conference on Neural Information Processing Systems}, 2024.
\newblock URL \url{https://openreview.net/forum?id=0og7nmvDbe}.

\bibitem[Stolfo et~al.(2025)Stolfo, Balachandran, Yousefi, Horvitz, and Nushi]{stolfo2025improving}
Alessandro Stolfo, Vidhisha Balachandran, Safoora Yousefi, Eric Horvitz, and Besmira Nushi.
\newblock Improving instruction-following in language models through activation steering.
\newblock In \emph{The Thirteenth International Conference on Learning Representations}, 2025.
\newblock URL \url{https://openreview.net/forum?id=wozhdnRCtw}.

\bibitem[Team(2025)]{qwq32b}
Qwen Team.
\newblock Qwq-32b: Embracing the power of reinforcement learning, March 2025.
\newblock URL \url{https://qwenlm.github.io/blog/qwq-32b/}.

\bibitem[Turner et~al.(2023)Turner, Thiergart, Leech, Udell, Vazquez, Mini, and MacDiarmid]{turner2023steering}
Alexander~Matt Turner, Lisa Thiergart, Gavin Leech, David Udell, Juan~J Vazquez, Ulisse Mini, and Monte MacDiarmid.
\newblock Steering language models with activation engineering.
\newblock \emph{arXiv preprint arXiv:2308.10248}, 2023.

\bibitem[Vaswani et~al.(2017)Vaswani, Shazeer, Parmar, Uszkoreit, Jones, Gomez, Kaiser, and Polosukhin]{vaswani2017attention}
Ashish Vaswani, Noam Shazeer, Niki Parmar, Jakob Uszkoreit, Llion Jones, Aidan~N Gomez, {\L}ukasz Kaiser, and Illia Polosukhin.
\newblock Attention is all you need.
\newblock \emph{Advances in neural information processing systems}, 30, 2017.

\bibitem[Wang et~al.(2025)Wang, Liu, Bi, Zhang, Li, Ma, He, Yu, Li, Fang, et~al.]{wang2025safety}
Cheng Wang, Yue Liu, Baolong Bi, Duzhen Zhang, Zhong-Zhi Li, Yingwei Ma, Yufei He, Shengju Yu, Xinfeng Li, Junfeng Fang, et~al.
\newblock Safety in large reasoning models: A survey.
\newblock \emph{arXiv preprint arXiv:2504.17704}, 2025.

\bibitem[Wang et~al.(2024)Wang, Ma, Zhang, Ni, Chandra, Guo, Ren, Arulraj, He, Jiang, et~al.]{wang2024mmlu}
Yubo Wang, Xueguang Ma, Ge~Zhang, Yuansheng Ni, Abhranil Chandra, Shiguang Guo, Weiming Ren, Aaran Arulraj, Xuan He, Ziyan Jiang, et~al.
\newblock Mmlu-pro: A more robust and challenging multi-task language understanding benchmark.
\newblock \emph{Advances in Neural Information Processing Systems}, 37:\penalty0 95266--95290, 2024.

\bibitem[Wei et~al.(2024)Wei, Huang, Huang, Xie, Qi, Xia, Mittal, Wang, and Henderson]{wei2024assessing}
Boyi Wei, Kaixuan Huang, Yangsibo Huang, Tinghao Xie, Xiangyu Qi, Mengzhou Xia, Prateek Mittal, Mengdi Wang, and Peter Henderson.
\newblock Assessing the brittleness of safety alignment via pruning and low-rank modifications.
\newblock In \emph{Forty-first International Conference on Machine Learning}, 2024.
\newblock URL \url{https://openreview.net/forum?id=K6xxnKN2gm}.

\bibitem[Wu et~al.(2024)Wu, Wang, Xiao, Peng, and Fu]{wu2024retrieval}
Wenhao Wu, Yizhong Wang, Guangxuan Xiao, Hao Peng, and Yao Fu.
\newblock Retrieval head mechanistically explains long-context factuality.
\newblock \emph{arXiv preprint arXiv:2404.15574}, 2024.

\bibitem[Xu et~al.(2024)Xu, Huang, Chen, and Wang]{xu2024uncovering}
Zhihao Xu, Ruixuan Huang, Changyu Chen, and Xiting Wang.
\newblock Uncovering safety risks of large language models through concept activation vector.
\newblock \emph{Advances in Neural Information Processing Systems}, 37:\penalty0 116743--116782, 2024.

\bibitem[Yan et~al.(2025)Yan, Xu, and He]{yan2025thinking}
Hanqi Yan, Hainiu Xu, and Yulan He.
\newblock Thinking hard, going misaligned: Emergent misalignment in llms.
\newblock \emph{arXiv preprint arXiv:2509.00544}, 2025.

\bibitem[Yang et~al.(2024)Yang, Yang, Zhang, Hui, Zheng, Yu, Li, Liu, Huang, Wei, Lin, Yang, Tu, Zhang, Yang, Yang, Zhou, Lin, Dang, Lu, Bao, Yang, Yu, Li, Xue, Zhang, Zhu, Men, Lin, Li, Tang, Xia, Ren, Ren, Fan, Su, Zhang, Wan, Liu, Cui, Zhang, and Qiu]{qwen2.5}
An~Yang, Baosong Yang, Beichen Zhang, Binyuan Hui, Bo~Zheng, Bowen Yu, Chengyuan Li, Dayiheng Liu, Fei Huang, Haoran Wei, Huan Lin, Jian Yang, Jianhong Tu, Jianwei Zhang, Jianxin Yang, Jiaxi Yang, Jingren Zhou, Junyang Lin, Kai Dang, Keming Lu, Keqin Bao, Kexin Yang, Le~Yu, Mei Li, Mingfeng Xue, Pei Zhang, Qin Zhu, Rui Men, Runji Lin, Tianhao Li, Tianyi Tang, Tingyu Xia, Xingzhang Ren, Xuancheng Ren, Yang Fan, Yang Su, Yichang Zhang, Yu~Wan, Yuqiong Liu, Zeyu Cui, Zhenru Zhang, and Zihan Qiu.
\newblock Qwen2.5 technical report.
\newblock \emph{arXiv preprint arXiv:2412.15115}, 2024.

\bibitem[Yang et~al.(2025)Yang, Li, Yang, Zhang, Hui, Zheng, Yu, Gao, Huang, Lv, et~al.]{yang2025qwen3}
An~Yang, Anfeng Li, Baosong Yang, Beichen Zhang, Binyuan Hui, Bo~Zheng, Bowen Yu, Chang Gao, Chengen Huang, Chenxu Lv, et~al.
\newblock Qwen3 technical report.
\newblock \emph{arXiv preprint arXiv:2505.09388}, 2025.

\bibitem[Yi et~al.(2025)Yi, Zheng, Wang, de~Melo, Wang, and He]{yi2025nlsr}
Xin Yi, Shunfan Zheng, Linlin Wang, Gerard de~Melo, Xiaoling Wang, and Liang He.
\newblock Nlsr: Neuron-level safety realignment of large language models against harmful fine-tuning.
\newblock In \emph{Proceedings of the AAAI Conference on Artificial Intelligence}, volume~39, pp.\  25706--25714, 2025.

\bibitem[Yin \& Steinhardt(2025)Yin and Steinhardt]{yin2025which}
Kayo Yin and Jacob Steinhardt.
\newblock Which attention heads matter for in-context learning?
\newblock In \emph{Forty-second International Conference on Machine Learning}, 2025.
\newblock URL \url{https://openreview.net/forum?id=C7XmEByCFv}.

\bibitem[Yin et~al.(2025)Yin, Leong, Zhang, Zhu, Yan, Zhang, He, Li, Wang, Zhang, and Yang]{yin2025constrain}
Qingyu Yin, Chak~Tou Leong, Hongbo Zhang, Minjun Zhu, Hanqi Yan, Qiang Zhang, Yulan He, Wenjie Li, Jun Wang, Yue Zhang, and Linyi Yang.
\newblock Constrain alignment with sparse autoencoders.
\newblock In \emph{Forty-second International Conference on Machine Learning}, 2025.
\newblock URL \url{https://openreview.net/forum?id=BCKSxOFX85}.

\bibitem[Yu et~al.(2025)Yu, Chen, Lin, Zhou, Zheng, Yang, Men, Gao, Li, Dang, Liu, and Liu]{yu2025groupsequencepolicy}
Bowen Yu, Xiong-Hui Chen, Junyang Lin, Jingren Zhou, Chujie Zheng, An~Yang, Rui Men, Chang Gao, Mingze Li, Kai Dang, Yuqiong Liu, and Shixuan Liu.
\newblock Group sequence policy optimization, 2025.
\newblock URL \url{https://arxiv.org/abs/2507.18071}.

\bibitem[Zaremba et~al.(2025)Zaremba, Nitishinskaya, Barak, Lin, Toyer, Yu, Dias, Wallace, Xiao, Heidecke, et~al.]{zaremba2025trading}
Wojciech Zaremba, Evgenia Nitishinskaya, Boaz Barak, Stephanie Lin, Sam Toyer, Yaodong Yu, Rachel Dias, Eric Wallace, Kai Xiao, Johannes Heidecke, et~al.
\newblock Trading inference-time compute for adversarial robustness.
\newblock \emph{arXiv preprint arXiv:2501.18841}, 2025.

\bibitem[Zhang \& Nanda(2023)Zhang and Nanda]{zhang2023towards}
Fred Zhang and Neel Nanda.
\newblock Towards best practices of activation patching in language models: Metrics and methods.
\newblock \emph{arXiv preprint arXiv:2309.16042}, 2023.

\bibitem[Zhang et~al.(2024)Zhang, Yao, Tian, Wang, Deng, Wang, Xi, Mao, Zhang, Ni, et~al.]{zhang2024comprehensive}
Ningyu Zhang, Yunzhi Yao, Bozhong Tian, Peng Wang, Shumin Deng, Mengru Wang, Zekun Xi, Shengyu Mao, Jintian Zhang, Yuansheng Ni, et~al.
\newblock A comprehensive study of knowledge editing for large language models.
\newblock \emph{arXiv preprint arXiv:2401.01286}, 2024.

\bibitem[Zhang et~al.(2025)Zhang, Zeng, Li, Huang, Deng, and Dong]{zhang2025realsafe}
Yichi Zhang, Zihao Zeng, Dongbai Li, Yao Huang, Zhijie Deng, and Yinpeng Dong.
\newblock Realsafe-r1: Safety-aligned deepseek-r1 without compromising reasoning capability.
\newblock \emph{arXiv preprint arXiv:2504.10081}, 2025.

\bibitem[Zhao et~al.(2025)Zhao, Zhang, Xie, Goyal, Kawaguchi, and Shieh]{zhao2025understanding}
Yiran Zhao, Wenxuan Zhang, Yuxi Xie, Anirudh Goyal, Kenji Kawaguchi, and Michael Shieh.
\newblock Understanding and enhancing safety mechanisms of {LLM}s via safety-specific neuron.
\newblock In \emph{The Thirteenth International Conference on Learning Representations}, 2025.
\newblock URL \url{https://openreview.net/forum?id=yR47RmND1m}.

\bibitem[Zheng et~al.(2025)Zheng, Zheng, Cao, Tan, Liu, Wang, Liu, Yang, Su, Zhu, et~al.]{zheng2025beyond}
Baihui Zheng, Boren Zheng, Kerui Cao, Yingshui Tan, Zhendong Liu, Weixun Wang, Jiaheng Liu, Jian Yang, Wenbo Su, Xiaoyong Zhu, et~al.
\newblock Beyond safe answers: A benchmark for evaluating true risk awareness in large reasoning models.
\newblock \emph{arXiv preprint arXiv:2505.19690}, 2025.

\bibitem[Zhou et~al.(2023)Zhou, Liu, Xu, Iyer, Sun, Mao, Ma, Efrat, Yu, Yu, et~al.]{zhou2023lima}
Chunting Zhou, Pengfei Liu, Puxin Xu, Srinivasan Iyer, Jiao Sun, Yuning Mao, Xuezhe Ma, Avia Efrat, Ping Yu, Lili Yu, et~al.
\newblock Lima: Less is more for alignment.
\newblock \emph{Advances in Neural Information Processing Systems}, 36:\penalty0 55006--55021, 2023.

\bibitem[Zhou et~al.(2025{\natexlab{a}})Zhou, Liu, Zhao, Jangam, Srinivasa, Liu, Song, and Wang]{zhou2025hidden}
Kaiwen Zhou, Chengzhi Liu, Xuandong Zhao, Shreedhar Jangam, Jayanth Srinivasa, Gaowen Liu, Dawn Song, and Xin~Eric Wang.
\newblock The hidden risks of large reasoning models: A safety assessment of r1.
\newblock \emph{arXiv preprint arXiv:2502.12659}, 2025{\natexlab{a}}.

\bibitem[Zhou et~al.(2025{\natexlab{b}})Zhou, Yu, Zhang, Xu, Huang, Wang, Liu, Fang, and Li]{zhou2025on}
Zhenhong Zhou, Haiyang Yu, Xinghua Zhang, Rongwu Xu, Fei Huang, Kun Wang, Yang Liu, Junfeng Fang, and Yongbin Li.
\newblock On the role of attention heads in large language model safety.
\newblock In \emph{The Thirteenth International Conference on Learning Representations}, 2025{\natexlab{b}}.
\newblock URL \url{https://openreview.net/forum?id=h0Ak8A5yqw}.

\bibitem[Zhu et~al.(2025)Zhu, Yan, Wang, Yin, and Sha]{zhu2025reasoning}
Junda Zhu, Lingyong Yan, Shuaiqiang Wang, Dawei Yin, and Lei Sha.
\newblock Reasoning-to-defend: Safety-aware reasoning can defend large language models from jailbreaking.
\newblock \emph{arXiv preprint arXiv:2502.12970}, 2025.

\bibitem[Zhu et~al.(2024)Zhu, Yang, Wei, Zhang, and Zhang]{zhu2024locking}
Minjun Zhu, Linyi Yang, Yifan Wei, Ningyu Zhang, and Yue Zhang.
\newblock Locking down the finetuned llms safety.
\newblock \emph{arXiv preprint arXiv:2410.10343}, 2024.

\bibitem[Zou et~al.(2023{\natexlab{a}})Zou, Phan, Chen, Campbell, Guo, Ren, Pan, Yin, Mazeika, Dombrowski, et~al.]{zou2023representation}
Andy Zou, Long Phan, Sarah Chen, James Campbell, Phillip Guo, Richard Ren, Alexander Pan, Xuwang Yin, Mantas Mazeika, Ann-Kathrin Dombrowski, et~al.
\newblock Representation engineering: A top-down approach to ai transparency.
\newblock \emph{arXiv preprint arXiv:2310.01405}, 2023{\natexlab{a}}.

\bibitem[Zou et~al.(2023{\natexlab{b}})Zou, Wang, Kolter, and Fredrikson]{zou2023universal}
Andy Zou, Zifan Wang, J.~Zico Kolter, and Matt Fredrikson.
\newblock Universal and transferable adversarial attacks on aligned language models, 2023{\natexlab{b}}.

\bibitem[Zou et~al.(2024)Zou, Phan, Wang, Duenas, Lin, Andriushchenko, Kolter, Fredrikson, and Hendrycks]{zou2024improving}
Andy Zou, Long Phan, Justin Wang, Derek Duenas, Maxwell Lin, Maksym Andriushchenko, J~Zico Kolter, Matt Fredrikson, and Dan Hendrycks.
\newblock Improving alignment and robustness with circuit breakers.
\newblock In \emph{The Thirty-eighth Annual Conference on Neural Information Processing Systems}, 2024.
\newblock URL \url{https://openreview.net/forum?id=IbIB8SBKFV}.

\end{thebibliography}
